\documentclass[twocolumn,lengthcheck,superscriptaddress]{revtex4-1}
\usepackage{amsmath,amsfonts,mathrsfs}
\usepackage{algorithm}
\usepackage{caption}
\usepackage{subcaption}
\usepackage{algpseudocode}
\usepackage{graphicx}
\usepackage{indentfirst}
\usepackage{titlesec}
\usepackage{hyperref}
\usepackage[utf8]{inputenc}
\usepackage{natbib}

\begin{document}

 \title{\textbf{Physicist's Journeys Through the AI World - A Topical Review} \\  \textit{There is no royal road to unsupervised learning}}
\author{Imad\ Alhousseini} 
\email{imad.alhousseini@gmail.com}
\affiliation{Lebanese University, Faculty of Sciences Branch 1, Hadath, Beirut, Lebanon.}
\author{Wissam\ Chemissany}
\email{wissamch@caltech.edu}
\affiliation{Institute for Quantum Information and Matter, California Institute of Technology,\\
1200 E California Blvd, Pasadena, CA 91125, USA.}
\author{Fatima\ Kleit}
\email{fatimasamikl@gmail.com}
\affiliation{Lebanese University, Faculty of Sciences Branch 1, Hadath, Beirut, Lebanon.}
\author{Aly\ Nasrallah}
\email{aly.nasrallah94@gmail.com}
\affiliation{Lebanese University, Faculty of Sciences Branch 1, Hadath, Beirut, Lebanon.}

\begin{abstract}
	Artificial Intelligence (AI), defined in its most simple form, is a technological tool that makes machines intelligent. Since learning is at the core of intelligence, machine learning poses itself as a core sub-field of AI. Then there comes a subclass of machine learning, known as deep learning, to address the limitations of their predecessors. AI has generally acquired its prominence over the past few years due to its considerable progress in various fields. AI has vastly invaded the realm of research. This has led physicists to attentively direct their research towards implementing AI tools. Their central aim has been to gain better understanding and enrich their intuition. This review article is meant to supplement the previously presented efforts to bridge the gap between AI and physics, and take a serious step forward to filter out the ``Babelian" clashes  brought about from such gabs. This necessitates first to have fundamental knowledge about common AI tools. To this end, the review's primary focus shall be on deep learning models called artificial neural networks. They are deep learning models which train themselves through different learning processes. It discusses also the concept of Markov decision processes. Finally, shortcut to the main goal, the review thoroughly examines how these neural networks are capable to construct a physical theory describing some observations without applying any previous physical knowledge.
\end{abstract}

\maketitle
\begin{figure}

\begin{subfigure}{1\textwidth}
        \includegraphics[width=17cm,height=11cm]{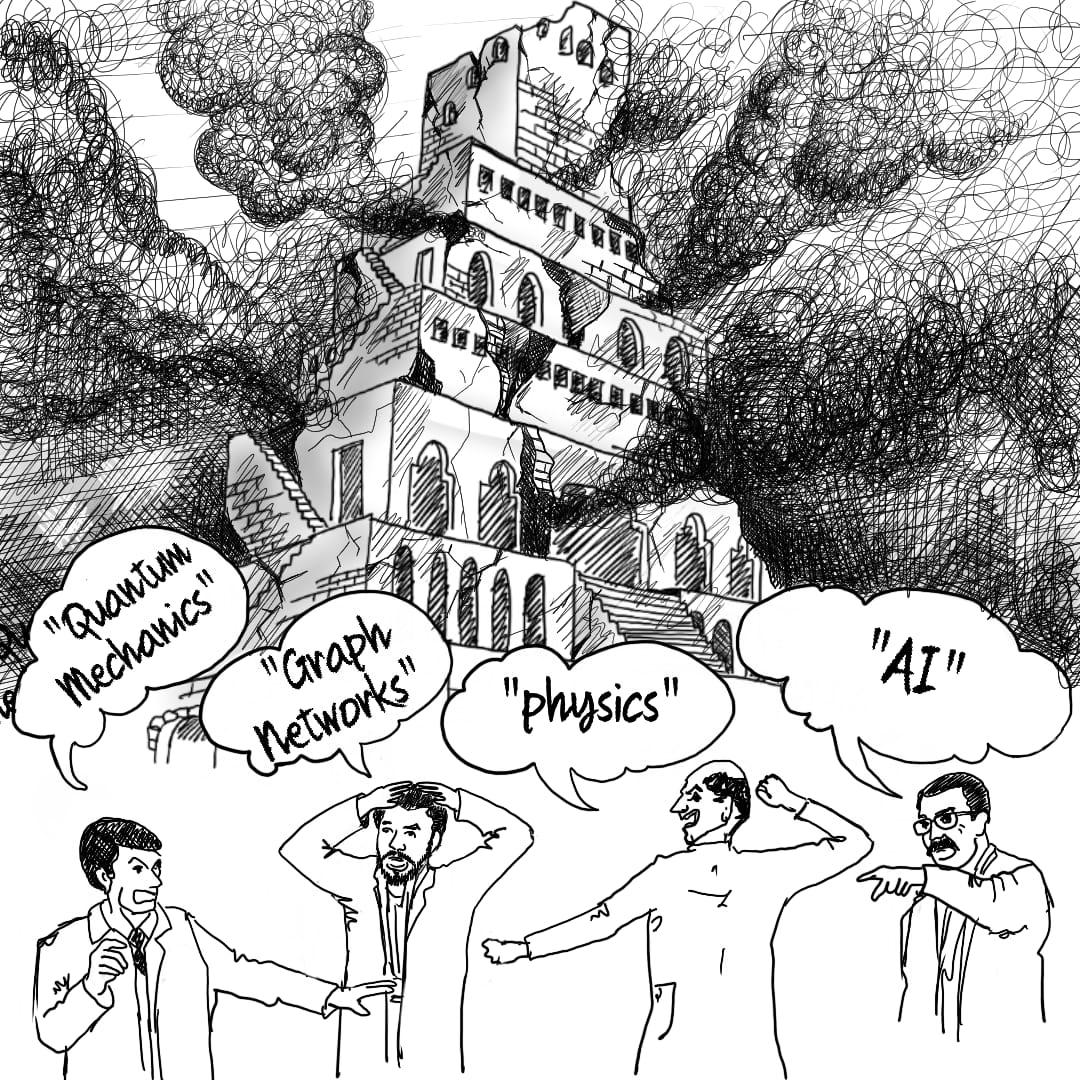} 
        \caption*{\textit{To reveal the secret of Babel, follow up!}}
    \end{subfigure}
\end{figure}
\clearpage
\tableofcontents

\section{INTRODUCTION}\label{ch1}
Artificial Intelligence (AI) is a broad field of science whose main objective is to make machines smarter. This means that machines are constructed so that they behave intelligently as humans do. Machines in that way are capable to adapt faster to whatever information they receive. AI acquired its prominence due to its considerable breakthroughs in various fields. Common real-life AI examples are self-driving cars, smartphones, and computer games. Based on this definition, this section defines the concept of machine learning, a sub-field of AI then introduces the concept of deep learning. After retailing these concepts thoroughly, this section highlights the relation between AI, in general, and physics. Finally, a brief summary discusses how the review is organized.


\subsection{Machine Learning: Cornerstone of AI}
Artificial Intelligence  is a broad field of science whose main objective is to make machines smarter. A fundamental subject of AI is machine learning (ML) \citep{feynman}. Machine learning implements the ability to learn from experience, i.e. observational data in hand. This is what makes machines intelligent since learning is at the core of intelligence. When a machine is fed with data, it first inspects it and extracts corresponding features (useful information). It then builds a model that is responsible for inferring new predictions based on those extracted features. Hence, the emphasis of machine learning is on constructing computer algorithms automatically without being explicitly programmed. This means that the computer will come up with its own program rather than having humans intervene in programming it directly.  Applications of ML techniques often create more accurate results in comparison to those of direct programming. ML meets with statistics, mathematics, physics, and theoretical computer science over a wide range of applications. Some of these real-life applications where ML is implemented include face detection, speech recognition, classification, medical diagnosis, prediction, and regression \cite{mehta2019high}.


\subsection{Machine Learning vs. Deep Learning}
Since their inception, ML techniques have achieved considerable success over direct programming. As discussed, one of the main tasks done by a machine learning model is to extract the features, but this task is very handy. If the number of features extracted is insufficient, then this will lead to predictions that are not accurate enough. The model is said to be highly biased. On other hand, if the number of features is more than enough to output predictions, the model will also be weak. It is thus said to be highly variant. For that, if the model fails to extract the features efficiently, careful engineering is necessary, i.e. an expert will intervene to make adjustments to improve the accuracy. This limits the scope of machine learning techniques.

\begin{figure}[hbt]
\begin{center}
	  \includegraphics[totalheight=5.5cm]{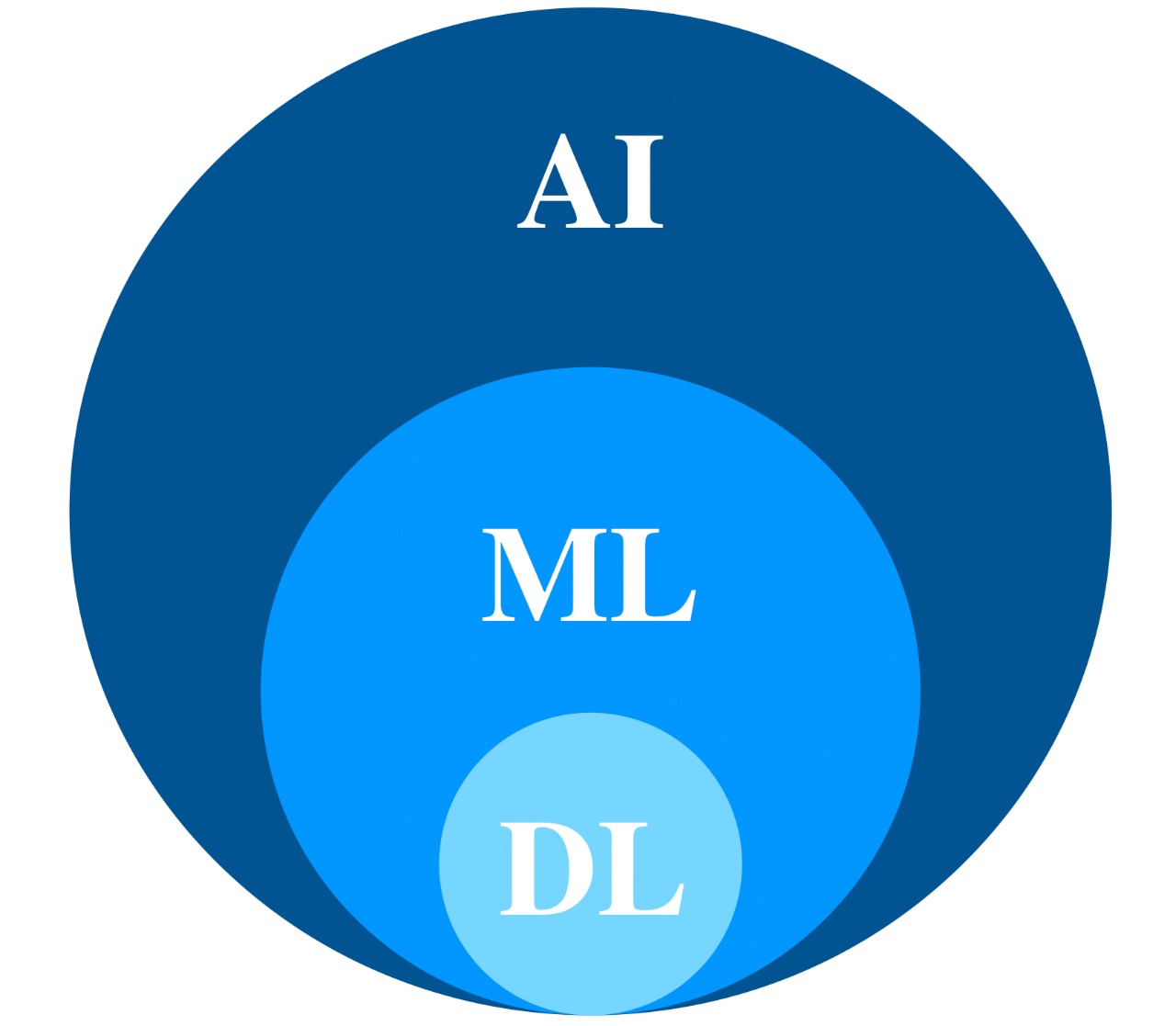}
  \caption{Artificial intelligence, machine learning and deep learning}
  \label{fig:DL}
\end{center}
\end{figure}

To address the aforementioned limitations, a new subset of ML emerged known as deep learning (DL). It is concerned with feature learning also known as representation learning Fig.(\ref{fig:DL}) which finds the features on its own from data where manual extraction of features isn't fully successful \citep{bengio2013representation}. Deep learning is implemented using complex architectures, often known as artificial neural networks (ANN's) mimicking the biological neural network of a human brain. For that, it is built in a logical structure to analyze data in a similar way a human draws conclusions. Upon analyzing the data, a neural network is able to extract features, make predictions, and determine how accurate the drawn conclusion is. In this way, deep learning model resembles the human intelligence.


\subsection{Physics and Machine Learning}
\subsubsection{Physics contributing to machine learning}
 Perhaps a question arises: What are the reasons behind unifying physics and machine learning? Upon going through the details of ML techniques, one of these reasons will be automatically manifested. It will be obvious then that the core concepts of ML techniques arise from the field of physics. Hence, physicists have been contributing to ML techniques since their early inception. Applying methods and theories developed in physics is still adopted in machine learning where efforts are present to explore new ML paradigms and develop physics-inspired learning algorithms. A group of researchers at Google, Princeton, Colombia, and MIT \citep{zeng2019tossingbot} confirmed this approach and designed a robot that develops an intuition of physics. No doubt that significant success has been made in improving the robots' efficiency in doing their tasks and learning from real world experiences. However, the researchers' understanding is that robots still need careful considerations. To address this challenge, they integrated simple physics models with deep learning techniques. Since physics explains how the real world works, this can be advantageous to support the robot with such models in a way to improve its capability to perform complex tasks. For example, to let the robot grasp objects efficiently, a neural network is provided with an image of the objects as an input in order to select the appropriate one from the box. At a certain stage, the network extracts the feature of the object, more specifically its position in the box. This feature along with the throwing velocity supplied by a physical simulator, are fed to another neural network. This network performs adjustments to predict a projectile that accurately targets the selected placing location. In conclusion, this unification between physics and deep learning techniques results in a better performance than techniques implemented alone.
 \subsubsection{Machine learning contributing to physics}
In turn, machine learning techniques can be used as a toolkit in physics. Physicists can benefit from ML when it comes to data analysis. Physics is one of the scientific fields that give rise to big data sets in diverse areas such as condensed matter physics, experimental particle physics, observational cosmology, and quantum computing. For example, the recent experiment "Event Horizon Telescope" recorded 5 petabytes of data in order to generate the first ever image of a super-massive black hole \citep{akiyama2019first}. That's why physicists are integrating ML techniques and following any advances in this direction. The benefits of machine learning for physicists don't stop here. Physicists implement different ML techniques with a view to improve their physical understanding and intuitions. To illustrate this approach, a recent work was done on neural networks to investigate whether they can be used to discover physical concepts even in domains that aren't clearly evident, such as quantum mechanics. This is done in the work of Renato \textit{et al.} in \citep{iten2018discovering} to be detailed in section(\ref{ch4}) as a first step in this approach. This depicts a promising research direction of how ML techniques can be applied to solve physical problems. The central question here: Can artificial intelligence discover new physics from raw data? This review will introduce to the reader attempts made recently as a first step to answer this question.


\subsection{Layout}
It must be emphasized that this review discusses how machine learning, and AI in general, interplay with physics. Since common AI tools are based on physical concepts, this is an indicator of how the AI community benefits from that of physics. However, this review highlights the other way around. Physicists are taking the challenge to make breakthroughs upon implementing AI tools in their research. Any approach in this direction appears to be very promising. To pave the way directly to the point, the review is organized as follows: section (\ref{ch2}) reviews fundamental concepts about artificial neural networks. These are a class of DL techniques that are self-trained through different learning processes: supervised, unsupervised, and reinforcement learning. Talking about reinforcement learning smooths the way to introduce the concept of Markov decision processes as explained in section (\ref{ch3}). Sections (\ref{ch4}) and (\ref{ch5}) fully retail two approaches to show how the DL techniques are used to help physicists improve their intuition about different physical settings. The former explains the strategy of how neural networks are implemented to describe physical settings, while the latter illustrates an algorithm that works the same way a physicist works upon dealing with a physical problem. Both, the algorithm and a physicist, use the four following strategies to solve any problem: divide-and-conquer, Occam's razor, unification, and lifelong learning. Finally, some concluding remarks are made with an opening to future work.


\section{ARTIFICIAL NEURAL NETWORKS} \label{ch2}
This section provides background knowledge on artificial neural networks. This knowledge is indispensable to understand following ML techniques through which they are implemented. This section first introduces the building block of ANNs: the artificial neuron, then discusses how information is being processed in ANNs. Here comes an important step to neural networks called training. The main objective of this step is that it leads neural networks to produce results with very high accuracy. This training occurs through an algorithm called gradient descent. All these topics are presented in the following sections.


\subsection{Artificial Neural Networks In a Nutshell}
\begin{figure}[hbt]
\begin{center}
	  \includegraphics[totalheight=4.8cm]{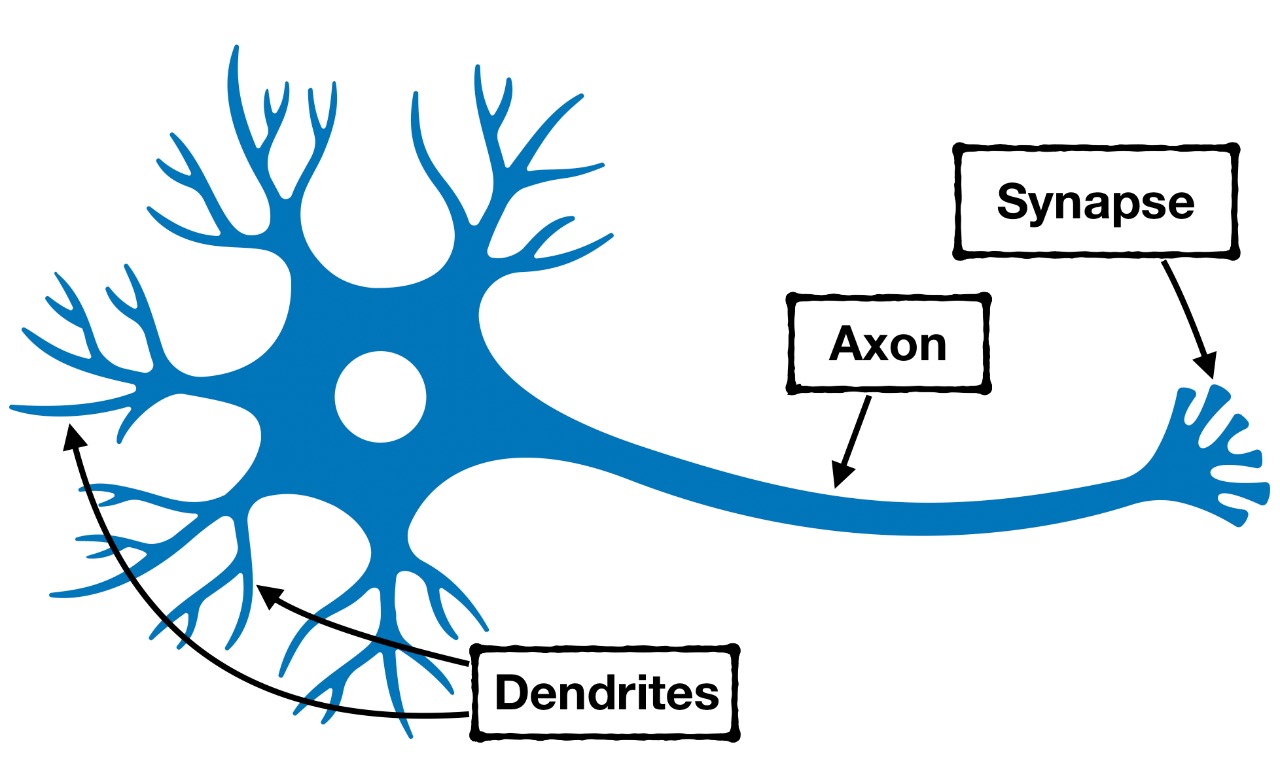}
  \caption{An illustration of a  biological neuron}
  \label{fig:neuron}
\end{center}
\end{figure}

An artificial neuron is a computational model that resembles a biological one \citep{haykin2009neural}. In the human body, electrical signals are transmitted among natural neurons through synapses located on the dendrites, i.e. membranes of the neuron. These signals activate the neuron whenever they exceed a specific threshold and therefore, a signal is emitted through the axon to activate the next neuron Fig.(\ref{fig:neuron}). Take for example the case when a human hand approaches a hot solid. If the solid is hot enough, the neurons will be quickly activated transmitting a command to warn off the hand. Otherwise, the human hand shows no reaction.

The artificial neuron with its basic components is analogous to the biological one; it is the building block of the artificial neural network \citep{kriesel2007brief}. ANN  consists of several interconnected consecutive layers where each layer is made up of stacked artificial neurons. The first layer is the input layer which receives the input data. This data is provided as a vector  \textbf{x} = $\{x_i\}_{i=1}^n$ where each neuron of the input layer is supplied with one element $x_i$. The inputs are multiplied by weights w = $\{w_i\}_{i=1}^n$ indicating the strength of each input, i.e. the higher the weight, the more influence the corresponding input has. The weighted sum of all inputs $\sum_{i=1}^n w_i x_i$ is then computed and an external bias, denoted by $b$, is added to it. The resulting value ($\sum_{i=1}^n [w_i x_i] + b$) is supplied as a variable to a mathematical function called the activation function $\psi(.)$. Its output is fed to a neuron in the next layer as an input. Examples of activation functions are presented in Appendix (\ref{activation}). The resulting computation any neuron receives basically depends on the incoming weights. It is important to note that the incoming weights into a specific neuron generally differ from those coming to any other neuron in the same layer. This results in a different computed input for each one. It is also worth mentioning that the bias is added to the weighted sum to modify the net input of the activation function. According to its sign, this net input is either increased or decreased. To make things clearer, consider the activation function to be a one-dimensional function $f(u)$ where $u = w_i x_i$. This function can be shifted by translation upon the addition of a constant $b$ to its parameter $u$, $f(u+b)$ . According to the sign of $b$, this function is shifted to the left or right allowing more flexibility in the choice of the value of the function thus affecting its output as well. The bias plays the role of this constant $b$ \citep{haykin2009neural}.

\begin{figure}[hbt]
\begin{center}
	  \includegraphics[totalheight=5cm]{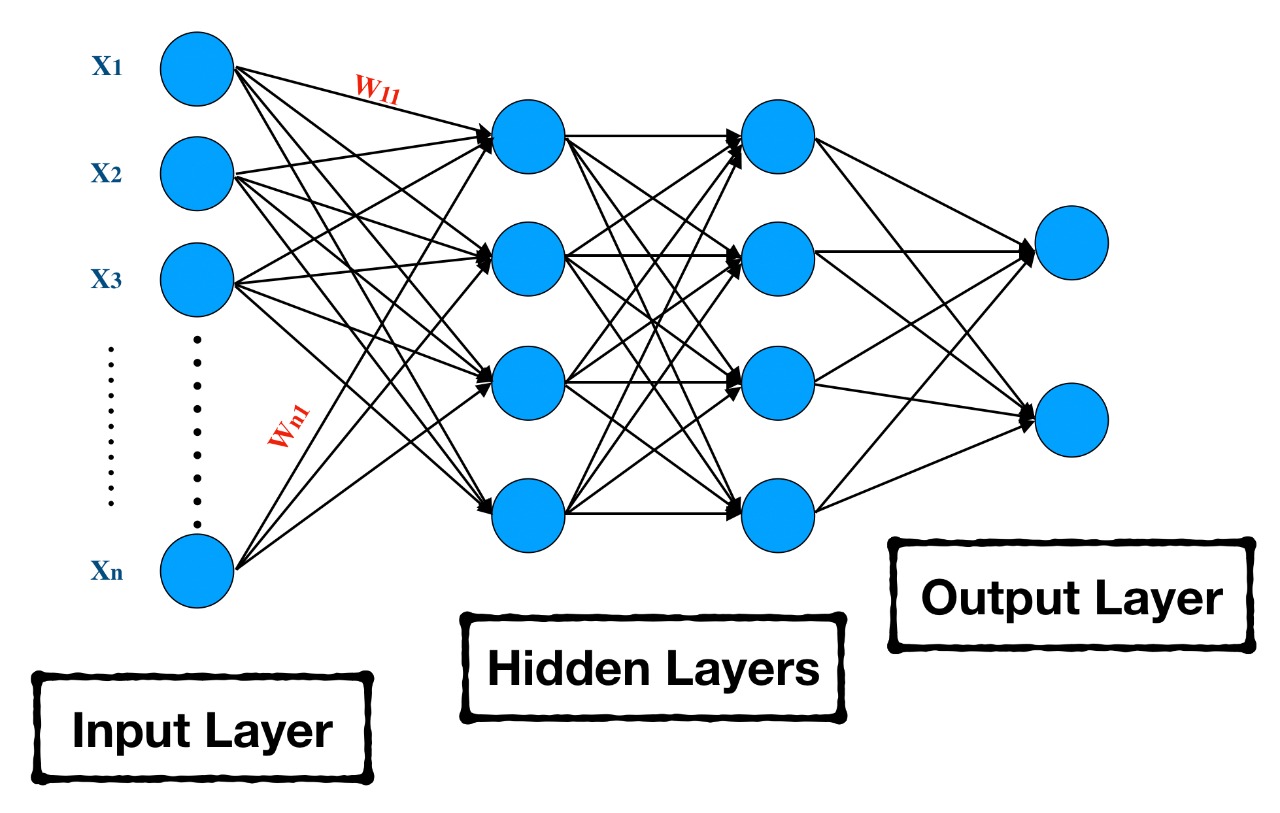}
  \caption{A model of an artificial neural network}
  \label{fig:ANN}
\end{center}
\end{figure}

The preceding steps are repeated along each layer of the neural network, thus information is being processed through it. Starting from the input layer and passing through intermediate layers known as hidden layers, the process ends with the output layer which holds the final desired results of the network Fig.(\ref{fig:ANN}). Perhaps, the simplest architecture of a neural network is that consisting of an input layer, a hidden layer with a sufficient number of hidden neurons, and an output layer \citep{mehta2019high}. This structure demonstrates the property of universality of neural networks which states that any continuous function can be approximated arbitrarily well by the aforementioned structure  of the neural network \citep{hornik1989multilayer,nielsen2015neural}.

More complex architectures are described as deep ones. This refers to neural networks that contain multiple hidden layers. Such structures are used frequently in modern research for their representational power due to the increased number of layers, and therefore the number of parameters, i.e. the weights and biases \citep{bengio2013representation}. Deep neural networks are able to learn more complex features from the input data. It is worth mentioning that the exact neural network architecture for a specific problem depends on several factors, two of which are: the type and amount of data that is available and the task to be achieved. Let's not forget also that the choice of the number of hidden layers and the number of hidden neurons in each layer alter the global performance of the network. In conclusion, a standard to abide by is that the number of parameters in a neural network should not be small enough to prevent under-fitting and not large enough to prevent over-fitting.


\subsection{Training}\label{train}
As mentioned in the previous section, the output of an artificial neuron depends on the adjustment of the parameters accompanied with the input data given to the neuron. However, in an artificial neural network composed of hundreds of interconnected neurons, all corresponding parameters cannot be set by hand. This is a very complicated task. Instead, regularizing the parameters of an artificial neural network occurs through a process often called training or learning. The parameters of a neural network hold a random initial value and then, after training, reach an optimal one. The optimization method is carried out with respect to a cost function that measures how close the output of a neural network is to the desired output of a specific input \citep{iten2018discovering}. This cost function must, in turn, be minimized and this minimization is performed through an algorithm often called gradient descent \citep{kriesel2007brief}. This algorithm is discussed thoroughly in the following subsection.

\subsubsection{Gradient descent}
The cost function $C_\theta$ is a multi-variable function that we aim to minimize during the learning process. It is as a function of the parameters $\theta$ of the neural network, and these parameters are adjusted iteratively and slowly until a minimum cost is achieved. 
Learning parameters using gradient descent takes the following steps:
\begin{enumerate}
\item The parameters of the neural net are randomly initialized $\theta_0$.
\item In each iteration $t$ and for each parameter $\theta_i$, the first-order gradient of the cost function $\nabla_{\theta_i} C_\theta$ is computed.
\item The parameter is then updated by
\begin{equation}
\theta_i = \theta_i - \eta\nabla_{\theta_i} C_\theta,
\end{equation}
where $\eta$ is the learning rate that is a hyper-parameter defining the step size of updating.
\item These steps are repeated for every iteration and the parameters are updated until the minimal cost is achieved.
\end{enumerate}
As seen, the term gradient descent corresponds then to decreasing the gradient step by step through adjusting the parameters until convergence. The choice of $\eta$ should be taken as not to take too big of a step leading to a collapse and not too small of a step leading to a slow performance.

\subsubsection{Stochastic Gradient Descent}
The cost function is often encountered as a summation of sub-functions, for example
\begin{equation}
C(\theta) = \frac{1}{n} \sum_{i=1}^{n} C_i(\theta),
\end{equation}
where $C_i(\theta)$ may be the euclidean distance between the desired output $y$ and the prediction $\hat{y}(\theta)$ i.e. $C_i= |y_i - \hat{y}_i(\theta)|$, and $n$ is the total number of input data points.

The gradient of the cost function with respect to a weight is then the sum of gradients of all $n$ sub-functions with respect to that weight. For a single step towards the minimum, the gradient will be calculated over the whole points. This is time-consuming especially if the number of data points is large. Stochastic gradient descent (SGD) \citep{nielsen2015neural} tackles this problem by taking only a subset $m$ of the data  at random to compute the gradient:
\begin{equation}
\nabla_{\theta_i} C(\theta_i) = \frac{1}{m} \sum_{j=1}^{m} \nabla_{\theta_i} C_j(\theta_i),
\end{equation}
where $m$ is termed mini-batch. Using this gradient, the parameter $\theta_i$ is then updated. The steps are repeated but with each iteration, the choice of the mini-batch must be changed.

\subsubsection{Adam}
One arising problem in gradient descent and stochastic gradient descent is the need to specify the learning rate. If the rate was too high, the algorithm may collapse. If it is too low, the performance will be slow. Standing for adaptive~moment~estimation, the Adam algorithm as introduced in \citep{kingma2015adam} takes a step towards solving this problem. The key idea of Adam algorithm is to compute a separate learning rate for each parameter of the neural network. It is an extension of the stochastic gradient descent method but with excellent results.

Briefly, Adam works as follows. For each parameter, the learning rate will be computed at each iteration. The algorithm for setting the learning rates starts by computing a hyper-parameter $m$ which is an estimate for the first moment that is the gradient (as seen in SGD). The update rule for $m$ is:
\begin{equation}
m_t = \beta _1 m_{t-1} + (1-\beta _1) \nabla _\theta L(\theta _{t-1}).
\end{equation}

Another hyper-parameter $v$ which is the estimate of the second moment that is the gradient  square is also computed at each iteration. The computation at each iteration is
\begin{equation} 
v_t = \beta _2 v_{t-1} + (1-\beta _2) (\nabla _\theta L(\theta _{t-1}))^2,
\end{equation} 
where $\beta_1$ and $\beta_2$ are factors empirically found to be 0.9 and 0.999 respectively. Since the first and second moments are initially set to 0, they remain close to 0 after each iteration especially that $\beta_1$ and $\beta_2$ are small. To solve this situation, a slight change is made 
\begin{equation}
\begin{aligned}
\hat{m}_t &= {m_t \over{1-\beta _1^t}},\\
\hat{v}_t &= {v_t \over{1-\beta _2^t}}.
\end{aligned}
\end{equation}
The update rule for the parameters at each iteration is then
\begin{equation}
\theta_t = \theta_{t-1} - \alpha {\hat{m}_t\over {\sqrt{\hat{v}_T} + \epsilon}},
\end{equation}
where $\alpha$ is the step size and $\epsilon = 10^{-8}$ is added to avoid any divergences. This procedure is repeated until convergence.

\subsection{Learning Paradigms} 
Hopefully, the previous sections have given a good overview of what an artificial neural network is and how it operates. As repeatedly mentioned, a neural network, like any artificial machine, thinks and behaves like a human. That's why it is trained to achieve such a goal. Upon training, a neural network is first supplied with a set of data called the training set. Then, it adjusts its parameters, as mentioned in section (\ref{train}), to continually learn from this data. Whenever the parameters reach their optimal values, training stops and in that way the neural network reaches the desired accuracy. The neural network is capable now to generalize and infer new predictions about data of the same type it did not encounter previously. Depending on the given training set, the processes through which a neural network learns differ. These are clarified in the following and can be easily generalized to any artificial machine \citep{haykin2009neural}.
\subsubsection{Supervised learning}
 Perhaps, supervised learning seems to be the simplest learning paradigm. An ANN is trained with a data set consisting of labeled data, i.e. data points augmented with labels. The neural network's role is to find a mapping between these pairs. In this way, upon linking each data point to its corresponding label, the data points are classified. When the finishes, the neural network employs this mapping to find labels to unseen data.
\subsubsection{Unsupervised learning}
 In contrast to the preceding paradigm, this one is given unlabeled data, i.e. the labels of the data are not provided. This necessitates that the neural network finds some relationships among the data in a way to cluster, i.e. group them. The grouping can be done either by categorizing or by ordering. A sufficiently trained neural network uses the inferred clustering rule and applies it on data it did not process previously.
\subsubsection{Reinforcement learning}
It is important to note that reinforcement learning differs from the previous paradigms. This paradigm essentially consists of a learning agent interacting with its environment. Hence, the environment in reinforcement learning plays the same role as the data in the previous paradigms. The process of learning in this case is evaluative: the learning agent receives a reward whenever it performs an action in the environment it is put in. Therefore, the goal of the agent is gaining the maximum possible reward. One approach to model the environment is to characterize it as Markov decision processes, i.e. the environment is defined as a set of states. Reinforcement Learning is discussed separately in the next section.


\section{REINFORCEMENT  LEARNING AND MARKOV DECISION PROCESSES }\label{ch3}
Machines, like human beings, are able to move around in an environment and interact either with it or with each other. However, their behaviors are not the same of course. Humans can interact adaptively and even intelligently when they encounter any environment including any stochastic behavior. However, these stochastic behaviors are troublesome for machines. Unlike previous attempts that directly engineer robots to accomplish specific tasks, now robots are made to behave independently without any human intervention.

The learning scheme for the robot is known as sequential decision making. The robot, or the agent as generally named, is left to take its own decisions sequentially in a series of time steps. So, the agent is the decision maker here and the learner as well. Definitely then, the agent performs actions and is rewarded based on the action performed at each step. In that way, the agent wanders the environment. The idea of the reward is to inform the agent of how good it is to take this action or how bad. The main goal is to increase the total rewards as much as possible.

Markov decision process (MDP) is a fundamental formalism that deals with the agent's interaction with the environment \citep{sutton1998introduction}. It assumes that the environment is accessible, i.e. the agent knows exactly where it is in it. This formalism models the environment as a set of states and the agent acts for improving its ability to behave optimally. It aims to figure out the best way to behave so that it achieves the required task in an optimal way. The agent's state is Markov that is it has all the sufficient information it needs to proceed; no need to check its history. The future is thus independent of past events.

The sequence of actions taken by the agent to reach the goal define the policy followed. The MDP framework allows learning an optimal policy that maximizes a long-term reward upon reaching a goal starting from an initial state. To address this challenging goal, we first introduce all the components of MDP, then we head to discuss the two algorithms that are used to compute the optimal behaviors: reinforcement learning and dynamic programming.


\subsection{Components of MDP}
Markov decision process \citep{sutton1998introduction} is the formalism defined as a tuple ($S,A,P,R$) where $S$ is a finite set of states, $A$ a finite set of actions, $P$ a transition function or probability and $R$ is a reward function.
\begin{enumerate}

\item States: As mentioned before, the environment is modelled as a set of finite states $S = \{s^1, ... , s^N\}$ where the size of the state space is $N$. The state is a unique characterization of all the features sufficient to describe the problem that is modelled. For example, the state in a chess game is a complete configuration of board pieces of both black and white.

\item Actions: The set of actions $A$ is defined as the finite set $\{a^1, ... , a^K\}$ where the size of the action space is $K$. Any action can be applied on any state to control it.

\item Transition Function: The transition function $P$ is defined as
\begin{equation*}
P=S\times A\times S\rightarrow[0,1].
\end{equation*}
$P$ defines a probability distribution over the set of all possible transitions, i.e. the conditional probability of changing from a current state $s \in S$ to a new state $s' \in S$ when applying an action $a \in A$. For all states $s$ and $s'$ and for all actions $a$, it is required that $0\leq P(s,a,s') \leq 1$. Furthermore, for all states $s$ and actions $a$, we have $\sum_{s' \in S} P(s,a,s') = 1$.
 Based on this and the fact that the system is Markovian, we can ensure that
\begin{align*}
&P(s_{t+1}|s_{t}, a_{t},s_{t-1}, a_{t-1},... )=\\
&P(s_{t+1}|s_{t},a_{t})=P(s_{t}, a_{t}, s_{t+1}).
\end{align*}

\item Reward Function: The state reward function is defined as
\begin{equation*}
R: S \rightarrow \mathbb{R}.
\end{equation*}
$R$ specifies a reward, i.e. a scalar feedback signal for being in a specific state after which an action is applied. This can be interpreted as negative (punishment) or positive (reward).
\end{enumerate}


\subsection{Policy and Optimality}
Given an MDP i.e knowing the set of states, actions, probabilities and rewards, a policy $\pi$ governs the action taken when present in a specific state. So the policy can be defined as 
\begin{equation*}
\pi: S \rightarrow A.
\end{equation*}
The policy thus controls the studied environment. There are two types of policies:
\begin{enumerate}
\item Deterministic policy that specifies the action $a$ taken in the state $s$: $\pi(s) = a$.
\item Stochastic policy that runs a probability distribution over the actions: $\pi(a|s)  = P(A_t = a|S_t=s)$. That is, it assigns probabilities to the actions that can be performed when present in state $s$.
\end{enumerate}

Under a certain policy $\pi$ and starting with a state $s_0$, the policy suggests an action $a_0$ to move to a state $s_1$. The agent receives a reward $r_0$ by making this transition. In this sense, the sequence under the policy is: 
\begin{equation*}
s_0,a_0,r_0,s_1,a_1,r_1,s_2,...
\end{equation*}

Our main goal is to find the optimal policy which is the policy that obtains the maximum number of rewards. It is important to note that the aim is not to maximize the immediate reward $R_t$, but rather the summation of all rewards collected during the task. These are expressed as a return function defined as:
\begin{equation}
G_t = R_{t+1} + R_{t+2}  ... + R_T,
\end{equation}
where $T$ is the final step. This makes sense when the task has limited steps; the return function will always converge. This model is known as finite-horizon. However, the interaction between the agent and the environment may be unlimited, and the agent may continue heading from one state to another and gathering rewards without achieving the goal. The return function will tend to infinity as more steps are being taken. The model of infinite-horizon is problematic. For this purpose, we introduce a discounting factor $\gamma$ which discounts the rewards, and the discounted return function is then:

\begin{equation}
\begin{aligned}
G_t &= R_{t+1} + \gamma R_{t+2} + \gamma^2 R_{t+2} ... \\
&= \sum\limits_{k=0}^{\alpha} \gamma^k R_{t+k+1},
\end{aligned}
\end{equation}

where $0\leq \gamma  \leq 1$. The discount factor $\gamma$ determines the importance of future rewards at the present. A reward after 2 steps is worth $\gamma^2$. It can be viewed as follows:
\begin{itemize}
\item if $\gamma =0$, then the agent is myopic and only cares about the immediate reward.
\item if $\gamma \rightarrow 0$, then the agent is near-sighted and cares about the nearest coming rewards.
\item if $\gamma \rightarrow 1$, then the agent is far-sighted and cares about future rewards.
\end{itemize}

The discount factor guarantees that the return function converges for a large number of steps. The return factor enjoys a recursive property
\begin{equation}
\begin{aligned}
G_t &= R_{t+1} + \gamma R_{t+2} + \gamma^2 R_{t+2} ... \\
&= R_{t+1} + \gamma G_{t+1}.
\end{aligned}
\end{equation}
The optimality criteria to maximize the return function depends on the problem at hand.


\subsection{Value Functions and Bellman Equations}
The value functions \citep{sutton1998introduction} of a state estimates how good it is to be present in this state in general or accompanied with taking a specific action. It depends on the future rewards to be gained starting from this state and following the policy. Value functions thus link optimality criteria to policies and are used to learn optimal policies.

A state-value function is the expected return when being present in that state under a particular policy:
\begin{equation}
\begin{aligned}
&v_{\pi}(s) = \mathbb{E}_{\pi} [G_t | S_t=s], \\
&= \mathbb{E}_{\pi} [ \sum\limits _{k=0}^{\alpha} \gamma^k R_{t+k+1} | S_t=s].
\end{aligned}
\end{equation}

A similar value function, denoted by $q(s,a)$, can be defined in the same way as the value of the state $s$, taking a specific action $a$ and thereafter following policy $\pi$:
\begin{equation}
\begin{aligned}
&q_{\pi}(s,a) = \mathbb{E}_{\pi} [G_t | S_t=s, A_t=a], \\
&= \mathbb{E}_{\pi} [ \sum\limits _{k=0}^{\alpha} \gamma^k R_{t+k+1} | S_t=s,A_t =a].
\end{aligned}
\end{equation}

The state-value functions satisfy a recursive relation:
\begin{equation}
\begin{aligned}
&v_{\pi}(s) = \mathbb{E}_{\pi} [G_t | S_t=s]
= \mathbb{E}_{\pi} [ R_{t+1} + \gamma G_{t+1} | S_t=s],\\
&= \sum_s \pi(a|s) \sum_{s'} \sum_{R} p(s,a,s')[R+ \gamma \mathbb{E}[G_{t+1} | S_{t+1}=s']],\\
&= \sum_s \pi(a|s) \sum_{s',R} p(s,a,s')[R+ \gamma v_{\pi}(s')].
\label{bellman}
\end{aligned}
\end{equation}

This equation is known as the Bellman Equation. It expresses the value function as the sum of all rewards and values of all possible future states weighted by their transition probabilities and a discount factor.
The optimal state-value function is thus:
 \begin{equation}
 v_* (s) = \text{max}~v_{\pi}(s),
 \end{equation}
and the optimal action-value function is:
\begin{equation}
q_{*} (s,a) = \text{max}~q_{\pi}(s,a).
\end{equation}

In the same manner, the optimal state-value function has a recursive property:

\begin{equation}
\begin{aligned}
v_{*} (s) = \text{max} \sum_{s}& \pi(a|s) \sum_{s',R} p(s,a,s')\\
&[R+ \gamma v_{*}(s')],
\label{optstate}
\end{aligned}
\end{equation}

\begin{equation}
\begin{aligned}
q_{*} (s,a) =  \sum_{s}& \pi(a|s) \sum_{s',R} p(s,a,s')\\
&[R+ \gamma \max_{a'}~q_{*}(s',a')].
\label{optact}
\end{aligned}
\end{equation}

This is known as the Bellman optimality equation. Finding $v_*$ or $q_*$ will be the corner stone in finding the optimal policy as will be seen in the following sections. To achieve the goal of arriving to the optimal policy, several algorithms have been proposed. these algorithms are divided in two classes: model-based and model-free algorithms. Both classes include the states and actions, but the model-based algorithms are also supplied with the transition probabilities and rewards, whereas the model-free aren't. In the following sections, these two cases are detailed by their corresponding algorithms; the first is dynamic programming which is model-based, the second is reinforcement learning which model-free.


\subsection{Dynamic Programming} 
Dynamic programming (DP) \citep{sutton1998introduction} is the category of algorithms that go after an optimal policy given that the dynamics of the environment (transition probabilities and rewards) are completely supplied. Dynamic programming is thus a model-based algorithm for solving MDPs.

\subsubsection{Reaching optimality: evaluation, improvement and iteration}

Finding the optimal policy of course follows from obtaining optimal value functions of the states: $v_{*}$ or $q_{*}$ which satisfy Bellman's optimality equations Eq.(\ref{optstate}, \ref{optact}). The general idea is that DP algorithms find the optimal value functions by updating the previous equations and then finding the optimal policy based on the value functions. The path of reaching the optimal policy thus mainly consists of two steps: evaluating then improving. Afterwards we repeat these steps several times till the optimum is achieved.

\underline{Policy evaluation}: We kick off by randomly considering some policy where the dynamics of the environment are completely known. We aim to find the state-value functions under this policy. These values satisfy Bellman's equation Eq.(\ref{bellman}). Solving this equation requires solving a system of $|\cal{S}|$ equations with $|\cal{S}|$ unknown value-state functions where $|\cal{S}|$ is the dimension of the state space, and this is tedious to achieve. One way to go around this problem is to transform it into an iterative problem:

\begin{enumerate}
\item We start with initializing the $v_\pi(s)$ for all states with arbitrary values, usually with zero.
\item Using Bellman equation, we evaluate all the value functions for all states.
\item Having evaluated the functions in the first round, we repeat the evaluation on and on such that the Bellman equation is now updated to: 
\begin{equation}
v_{k+1}(s) = \sum _{a} \pi(a,s) \sum _{s'}p(s',a,s)[R(s,a,s')+\gamma v_{k}(s')],
\end{equation}
where $k$ represents the iteration. This means that the value of the state in the current iteration depends on the value of the successor states in the previous iterations.
\item We continue updating the values of states by iterating until the current value doesn't differ much from the previous value i.e. :
\begin{equation}
|v_{k+1}(s) - v_{k}(s)| < \epsilon.
\end{equation}
\end{enumerate}
This is known as iterative~evaluation~policy. The final value obtained for each state $s$ under the given policy $\pi$ is then $v_{\pi}(s)$.

\underline{Policy improvement:} After computing all the state-value functions under a certain policy, we need to know whether being in this state $s$ and performing the action governed by the policy $\pi$ is better or worse than performing an action $a$ governed by some other policy $\pi'$. In other words, once in state $s$, we perform an action $a=\pi'(s) \neq \pi(s)$ and then continue with policy $\pi$. Is this better or worse?

This is answered using the state-action value function $q_{\pi}(s,a)$ where $a=\pi'(s)$ :
\begin{equation}
\begin{aligned}
q_{\pi}(s,a) = \sum_{s'}p(s',a,s)[R(s,a,s')+\gamma~v_{\pi}(s')].
\end{aligned}
\end{equation}

If $q_{\pi}(s,a)$ is in fact better than $v_{\pi}(s)$, then choosing this action $a=\pi'(s)$ then following $\pi$ is better than considering just $\pi$ from the beginning. The new policy $\pi'$ is thus an improved policy. This is know as policy~improvement~theorem. Having two deterministic policies $\pi$ and $\pi'$ and
\begin{equation}
q_{\pi}(s,\pi'(s)) \geq v_{\pi}(s),
\end{equation}
is indeed the same as having 
\begin{equation}
v_{\pi'}(s) \geq v_{\pi}(s).
\end{equation}

It is logical to sweep over all states present and the actions assigned to each of them and choose which action increases the value of each state according to $q_{\pi}(s,a)$.

The policy that aims to choose the action that increases the value of a certain state is known as the greedy~policy where
\begin{equation}
\begin{aligned}
\pi'(s) &= \arg \max _{a} ~q_{\pi}(s,a), \\
&= \arg \max _{a}~\mathbb{E}~[R_{t+1} + \gamma v_{\pi}(s')],\\
&= \arg \max _{a}~\sum_{s'} p(s,a,s')[R(s,a,s') + \gamma v_{\pi}(s')],
\end{aligned}
\end{equation}
where $\arg \max_{a}$ is the action that maximizes the action-state value function $q_{\pi}(s,a)$.
Therefore, the greedy policy is the policy that improves the value of the state by choosing a better action; this process of obtaining a greedy policy is the process of policy~improvement. If $v_{\pi} = v_{\pi'} $, then both $\pi$ and $\pi'$ are the optimal policies.

Note that if there are several actions that maximize the value function, then these actions must all be considered and given certain probabilities. This is the case where the policies aren't deterministic but rather stochastic.

\begin{figure}[hbt]
\centering
     \begin{subfigure}[b]{0.4\textwidth}
         \centering
         \includegraphics[width=\textwidth]{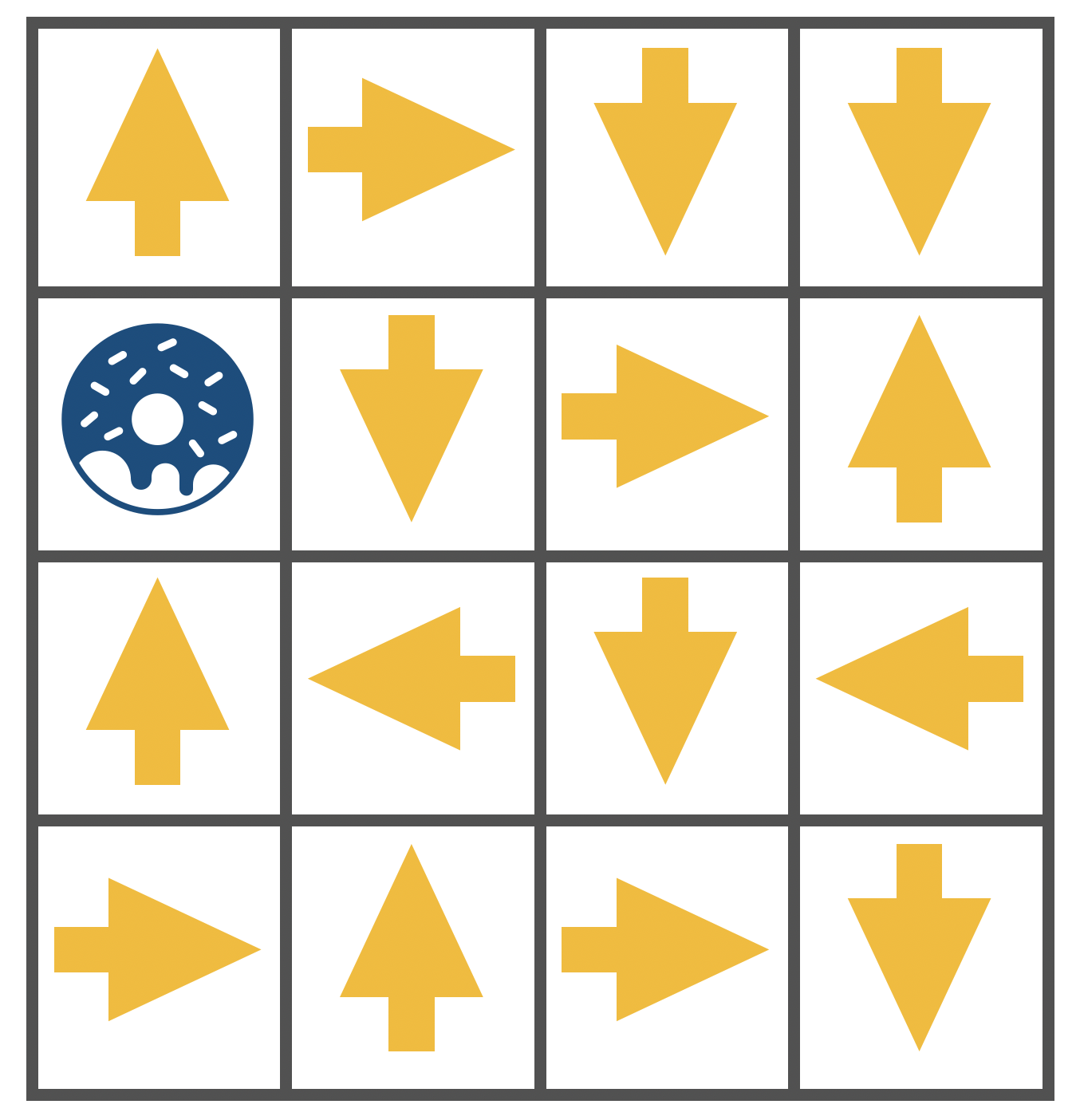}
         \caption{Policy randomly initialized. The goal is to get the yummy donut.}
         \label{fig:mdp1}
     \end{subfigure}
  \hfill
\begin{subfigure}[b]{0.395\textwidth}
         \centering
         \includegraphics[width=\textwidth]{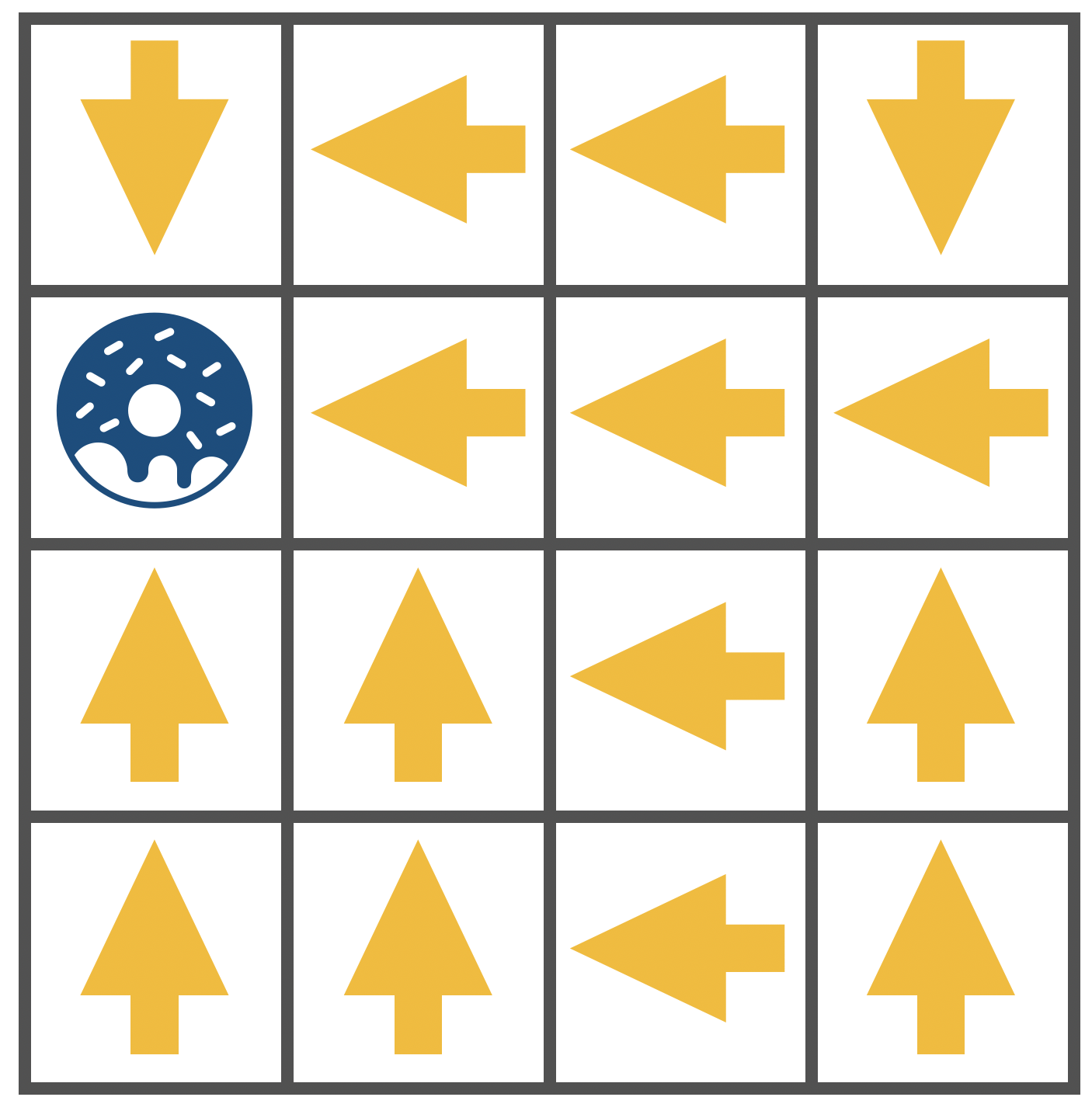}
         \caption{Optimal policy retrieved. After enough iterations, the agent eats the donut.}
         \label{fig:md2}
     \end{subfigure}
\end{figure}

\underline{Policy iteration:} Our main goal is to obtain the optimal policy, and as we've mentioned before it is the process repeating two steps successively: policy evaluation and policy improvement. This is policy~iteration. Starting with a policy $\pi$, we evaluate this policy, then we improve it to get the policy $\pi'$. By evaluating $\pi'$ then improving it, we'll end up with $\pi''$. Repeating the same process again and again, it'll converge to an optimal policy $\pi^*$ which is our goal. An example on how to start with a random policy and end up with an optimal one is illustrated in Fig.(\ref{fig:mdp1}, \ref{fig:md2}).

\subsubsection{Value iteration}
Policy iteration is costly since it requires sweeping over the set of states several times during the policy evaluation step. Can this step be reduced to include just one step i.e. can we substitute the iterations to obtain the value with only one step?

This is the process of value~iteration. Instead of sweeping the whole set of states several times to obtain the value then looking for the best action performed, we immediately do this in one step using an update for the Bellman equation:
\begin{equation}
v_{k+1}(s) = \max _{a} \sum_{s'} p(s,a,s')[R(s,a,s')+\gamma v_{k}(s')].
\end{equation}

Policy evaluation is still present, but it requires to take the action that maximizes the value. Thus value iteration joins policy evaluation and improvement in one making the convergence to the optimal policy faster. It is important to mention that some sweeps use value iteration while others still use policy evaluation, but the end result is always an optimal policy. 

\subsubsection{Asynchronous dynamic programming}
As discussed, sweeping over the large set of states is very costly even for just one sweep. Asynchronous DP doesn't sweep over the whole set of states but rather just over a subset in each sweep. A value of one state is updated using whatever values of the other states are available, one state can be updated several times whereas another state can be updated just once or twice. Asynchronous DP allows flexibility in choosing what states will be updated in this step and what states will remain the same under the condition that at the end of the whole process, every state must have been updated and not completely ignored. Some states require frequent updates whereas others require updating every now and then; some states are irrelevant to the process of reaching optimality and could be ignored all along.

\subsubsection{Generalized policy iteration}
In the preceding sections we saw how policy iteration lead us to find the optimal policy. It consists of two steps: policy evaluation and policy improvement. One step doesn't start unless the previous has terminated. Of course other processes are present to make policy iteration more efficient such as value iteration and asynchronous dynamic programming.

Generalized~policy~iteration (GPI) describes the process of policy evaluation and policy improvement whether the other processes are present or not. The whole idea as previously explained is that the current policy is evaluated then we improve the policy according to a better value function. Improvement and evaluation are thus interacting, and one drives the other. All model-based and model-free algorithms depend on GPI. Once the value function and improvement produce no change, then the optimal policy is reached.


\subsection{Reinforcement Learning}
The previous section discussed dynamic programming which is a model-based algorithm assuming that all the transition and reward functions are given to compute the optimal policy. When such a model is not available, reinforcement learning steps in. It necessitates statistical knowledge of the unknown model in a way to generate samples of state transitions and rewards. Sampling occurs due to the agent's interaction with the environment by doing actions to learn the optimal policy by trial-and-error. An important aspect must be highlighted then, namely the need for the agent's exploration of the environment. The agent must always try to perform different actions seeking better ones and not only exploit its current knowledge about good actions. Several strategies for exploration can be abided by. The most basic one is known as the $\epsilon$-greedy policy. The agent through its exploration chooses its current best action with a probability $1-\epsilon$ and any other action is taken randomly with probability $\epsilon$. This is a  reinforcement learning technique.

Reinforcement learning can be solved indirectly. This occurs upon the interaction with the environment by learning the transition and rewards functions and building up an approximate model of the MDP. Hence, all the dynamics, i.e. state values and state-action values, of the system can be deduced using all the methods of DP mentioned in the previously. Another option suggests estimating directly the state values of the actions without even estimating a model of the MDP. This option is known as direct Reinforcement Learning. Indeed, this happens to be a choice taken in model-free contexts. There exists other choices including temporal difference learning, Q-learning \citep{sutton1998introduction} and SARSA (State-Action-Reward-State-Action) \citep{graepel04learning}.

We detail in the next section the trust region policy optimization (TRPO) algorithm which is an algorithm using reinforcement learning. It improves the policy iteratively with cautious step sizes. An example of using TRPO efficiently will follow the algorithm.


\subsubsection{Trust region policy optimization}
Another method to reach the optimal policy is by using the Trust Region Policy Optimization algorithm \citep{schulman2015trust}. This algorithm outdoes other policy improvement algorithms due to the fact that it specifies a trust region for the step to be taken for improving the policy i.e. it takes the largest possible trusty step. In general, taking large steps is very risky and taking small steps makes the process very slow. TRPO solves this problem by defining a trust region to take the best steps avoiding a collapse of the improvement process.

As explained in \citep{schulman2015trust}, the procedure will start off by monotonically improving the policy through minimizing a certain loss function, then introducing approximations that are the core of the practical TRPO algorithm.

Considering an infinite-horizon MDP $(S,A,P,r,\rho_0,\gamma)$ where $\rho_0$ is the probability distribution of the initial states $s_0$. Recall the functions defined for an MDP:
\begin{itemize}
\item The state-action value function: 
\begin{equation} 
Q_{\pi} (s_t,a_t) =  \mathbb{E}_{s_{t+1}, a_{t+1}, ..} \Big[ \sum _{l=0}^{\infty} \gamma^{l} r(s_{t+l})\Big],
\end{equation}
\item the value function:
\begin{equation} 
V_{\pi} (s_t) =  \mathbb{E}_{a_{t},s_{t+1}, a_{t+1}, ..} \Big[ \sum _{l=0}^{\infty} \gamma^{l} r(s_{t+l})\Big],
\end{equation}
\item and the expected rewards:
\begin{equation} 
\eta (\pi) =  \mathbb{E}_{a_{0},s_{0},..} \Big[ \sum _{t=0}^{\infty} \gamma^{t} r(s_{t})\Big].
\end{equation}
\end{itemize}
A new function that quantifies how well the action performs compared to the average actions is defined as
\begin{itemize}
\item The advantage function:
\begin{equation} 
A_{\pi} (s, a) = Q_{\pi}(s, a) - V_{\pi}(s).
\end{equation}
\end{itemize}
Given two stochastic policies $\pi$ and $\tilde{\pi}$ where $\pi: S\times A \rightarrow [0,1]$, we can prove the expected rewards following the policy $\tilde{\pi}$ as:
\begin{equation}
\eta_{\tilde{\pi}} = \eta_{\pi} + \mathbb{E}_{\tilde{\pi}} \Big[ \sum _{t=0}^{\infty} \gamma^{t} A_{\pi} (s_t,a_t) \Big ].
\label{eqn: policyreward}
\end{equation}
Let's prove this formula as done in \citep{kakade2002approximately}. Start by the advantage of $\tilde{\pi}$ over $\pi$:
\begin{equation}
\begin{aligned}
&\mathbb{E}_{\tilde{\pi}} \Big[ \sum _{t=0}^{\infty} \gamma^{t} A_{\pi} (s_t,a_t) \Big ] 
=~\mathbb{E}_{\tilde{\pi}} \Big[ \sum_{t=0}^{\infty} \gamma^{t} \big (Q_{\pi} (s_t,a_t) - V_{\pi} (s_t) \big) \Big ],\\
=&~\mathbb{E}_{\tilde{\pi}} \Big[ \sum _{t=0}^{\infty} \gamma^{t} \big (R (s_t,a_t,s_{t+1}) + \gamma V_{\pi} (s_{t+1}) - V_{\pi} (s_t) \big) \Big ],\\
=&~\mathbb{E}_{\tilde{\pi}} \Big[ \sum _{t=0}^{\infty} \gamma^{t} R (s_t,a_t,s_{t+1}) \Big] + \mathbb{E}_{\tilde{\pi}} \Big[ \sum _{t=0}^{\infty} \big(\gamma^{t+1} V_{\pi} (s_{t+1}) - \gamma^{t}V_{\pi} (s_t) \big)\Big],\\
=&~\eta(\tilde{\pi}) + \mathbb{E}_{\tilde{\pi}} \Big[ \sum _{t=1}^{\infty}\gamma^{t} V_{\pi} (s_{t}) - \sum _{t=0}^{\infty}\gamma^{t}V_{\pi} (s_t)\Big],\\
=&~\eta(\tilde{\pi}) + \mathbb{E}_{\tilde{\pi}} \Big[ -V_{\pi} (s_{0})\Big],\\
=&~\eta(\tilde{\pi}) -\eta({\pi}).
\end{aligned}
\end{equation}
The result will then be as in Eq.(\ref{eqn: policyreward}). Before continuing with the explanation, let's head briefly to talk about visitation frequencies \citep{si2004handbook}. The state visitation frequency is the distribution of the probability of passing through a certain state following a specific policy. It is thus defined for a state $s$ as: 
\begin{equation}
\rho_{\pi}(s) = P(s_0 =s) + \gamma P(s_1 =s) + \gamma^2P(s_2 =s) + \dots 
\end{equation}
where the first term is the probability of encountering the state $s$ at the first time step, the second term is the probability at the second time step and so on. It must be kept in mind that visitation probabilities are heavily changed with the change of policy.\newline
Starting once more with the advantage:
\begin{equation}
\begin{aligned}
&\mathbb{E}_{\tilde{\pi}} \Big[ \sum _{t=0}^{\infty} \gamma^{t} A_{\pi} (s_t,a_t) \Big ], \\
=& \sum _{t=0}^{\infty} \sum_{s} P(s_t = s | \tilde{\pi}) \sum_{a} \tilde{\pi}(a|s) \gamma^{t} A_{\pi}(s,a),\\
=& \sum_{s} \sum _{t=0}^{\infty}  \gamma^{t} P(s_t = s | \tilde{\pi}) \sum_{a} \tilde{\pi}(a|s)  A_{\pi}(s,a),\\
=& \sum_{s} \rho_{\tilde{\pi}}(s) \sum_{a} \tilde{\pi}(a|s)  A_{\pi}(s,a).
\end{aligned}
\end{equation}
Combining this result with Eq.(\ref{eqn: policyreward}), we end up with
\begin{equation}
\eta(\tilde{\pi}) = \eta(\pi) + \sum_{s} \rho_{\tilde{\pi}}(s) \sum_{a} \tilde{\pi}(a|s)  A_{\pi}(s,a).
\label{eqn:opt}
\end{equation}
This equation implies that having a non-negative sum of expected advantage functions will increase the expected rewards when updating from policy $\pi$ to policy $\tilde{\pi}$ thus making $\tilde{\pi}$ an improved policy. If the summation of the expected advantages is zero, then the optimal policy is now reached and the performance is now constant $\big(\eta(\pi) = \eta(\tilde{\pi})\big)$.

As have mentioned before, the policy could be a deterministic policy such that $\tilde{\pi}(a|s) = 1$, and so improvement is guaranteed if at least one advantage function is positive with an existing visitation probability. However, if the policy is a stochastic one and the regime is an approximated regime, then due to the inevitable estimations error, there could be negative advantage functions. Moreover, the dependence of the visitation probability on the policy $\tilde{\pi}$ makes it really tedious to solve the optimization Eq.(\ref{eqn:opt}). For that, it is quite easier to use $\rho_\pi$ instead of $\rho_{\tilde{\pi}}$ in the optimization equation. This substitution is valid if the update from $\pi$ on $\tilde{\pi}$ is in a way that the changes in the visitation frequencies can be ignored. Then instead of Eq.(\ref{eqn:opt}), use 
\begin{equation}
L_{\pi}(\tilde{\pi}) = \eta(\pi) + \sum_{s} \rho_{\pi}(s) \sum_{a} \tilde{\pi}(a|s)  A_{\pi}(s,a).
\end{equation}
If the policy is parameterized and differentiable by a parameter $\theta$, then for the current policy $\pi_{\theta_{0}}$, there is
\begin{equation}
\begin{cases}
\begin{aligned}
L_{\pi_{\theta_0}}(\pi_{\theta_0}) &= \eta(\pi_{\theta_0}),\\
\nabla_\theta L_{\pi_{\theta_0}}(\pi_{\theta_0})|_{\theta = \theta_0} &= \nabla_\theta \eta(\pi_{\theta})|_{\theta = \theta_0}.
\end{aligned}
\end{cases}
\end{equation} 
This equation shows that any improvement from $\pi_{\theta_0}$ to $\tilde{\pi}$ which increases $L$ will definitely increase $\eta_{\tilde{\pi}}$, but it doesn't specify how good a step is. Recall that it's quite risky to take large steps and very slow to take small ones, so we must specify how big of step to take.

In the work of \citep{kakade2002approximately}, this issue was solved by defining the following lower bound
\begin{equation}
\eta(\tilde{\pi}) \geq L_{\pi}(\tilde{\pi}) - {{2\epsilon\gamma}\over{(1-\gamma)^2}}\alpha^2.
\end{equation}
This guarantees that increasing the right-hand side will surely increase the expected rewards under policy $\tilde{\pi}$ thus improving the policy. To tackle specifically stochastic policies, $\alpha^2$ will be the distance measure between the two policies such as the KL-divergence, more specifically, the maximum KL-divergence between the two policies $D_{KL}^{\max}(\pi,\tilde{\pi}) = {\max}_{s} ~D_{KL}(\pi(.|s) \| \tilde{\pi}(.|s))$ is taken to lower the bound further. Using the $D_{KL}^{\max}$, the bound becomes
\begin{equation}
\eta(\tilde{\pi}) \geq L_{\pi}(\tilde{\pi}) - C D_{KL}^{\max}(\pi,\tilde{\pi}).
\end{equation}
where $\epsilon = {\max}_{s,a} |A(s,a)|$ and $C={{4\epsilon\gamma}\over{(1-\gamma)^2}}$.
For the sake of simplicity, define the surrogate function $M_i(\pi_{i+1})$ such that 
\begin{equation}
M_i (\pi_{i+1}) = L_{\pi_i}(\pi_{i+1}) - C D_{KL}^{\max}(\pi_i,\pi_{i+1}).
\end{equation}
where $\pi_i$ is the current policy and $\pi_{i+1}$ is the new one. So,
\begin{equation}
\begin{aligned}
&\eta(\pi_{i+1}) \geq M_i (\pi_{i+1}), \\
\text{and,}~ &\eta(\pi_i) = M_i (\pi_i),\\
\text{then,}~ &\eta(\pi_{i+1}) - \eta(\pi_i) \geq M_i (\pi_{i+1}) - M_i (\pi_i).
\end{aligned}
\end{equation}
By maximizing the surrogate function $M_i(\pi_{i+1})$, it is guaranteed to have a monotonically increasing improvement of the policies i.e. $\eta_0 \leq \eta_1 \leq \eta_2 \leq \dots$ until optimization is reached. This algorithm is called the Minorization-Maximization algorithm where minorization corresponds to the fact that $\eta_i = M_i(\pi_i)$ is the lower bound, and maximization is quite obvious.

But we still haven't specified how big of a step to take. To do that, the trust region policy optimization algorithm \citep{schulman2015trust} is now presented as a practical approximation to the theoretical Minorization-Maximization algorithm.

Recall that the policies may be parameterized by some arbitrary parameter $\theta$. To make things a bit simpler, define the following notations as used by the authors in \citep{schulman2015trust}:
\begin{equation}
\begin{aligned}
\eta(\theta) &:= \eta{\pi_\theta}, \\
L_\theta(\tilde{\theta}) &:= L_{\pi_\theta}(\pi_{\tilde{\theta}}),\\
D_{KL}(\theta \| \tilde{\theta}) &:= D_{KL}(\pi_\theta \| \pi_{\tilde{\theta}}).
\end{aligned}
\end{equation}
And by denoting $\theta_{\text{old}}$ as the old parameters to be improved, the optimization equation becomes
\begin{equation}
\eta(\theta) \geq L_{\theta_{\text{old}}}(\theta) - C D_{KL}^{\max}(\theta_{\text{old}},\theta).
\end{equation}
The main goal can then be summarized by
\begin{equation}
\underset{\theta}{\text{maximize}}~ \big[L_{\theta_{\text{old}}}(\theta) - C D_{KL}^{\max}(\theta_{\text{old}},\theta)\big].
\label{eqn:surrogate}
\end{equation}
However, taking $C={{4\epsilon\gamma}\over{(1-\gamma)^2}}$ leads to small steps thus a slow rate of improvement. A constraint must be put on the KL-divergence in order to take larger steps but not to large as to cause collapses. That being said, this constraint is the $trust~region$ constraint on KL-divergence and the condition to satisfy now becomes:
\begin{equation}
\begin{aligned}
&\underset{\theta}{\text{maximize}}~ L_{\theta_{\text{old}}}(\theta),\\
&\text{subjected to} ~D_{KL}^{\max}(\theta_{\text{old}},\theta) \leq \delta.
\label{eqn:condition}
\end{aligned}
\end{equation}
The way of writing Eq.(\ref{eqn:condition}) is called Lagrangian duality where the constraint may be integrated back to the condition using a multiplier. The constraint put implies that each KL-divergence is bounded, but this tedious to work with due to the big number of constraints. A way to avoid this problem is by taking the average of the KL-divergence. The condition thus becomes: 
\begin{equation}
\begin{aligned}
&\underset{\theta}{\text{maximize}}~ L_{\theta_{\text{old}}}(\theta),\\
&\text{subjected to} ~\overline{D}_{KL}^{\rho_{\theta_{\text{old}}}}(\theta_{\text{old}},\theta) \leq \delta,
\label{eqn:conditionav}
\end{aligned}
\end{equation}
where $\overline{D}_{KL}^{\rho_{\theta}}(\theta_1,\theta_2) :=  \mathbb{E}_{\rho} \Big [D_{KL} \big( \pi_{\theta_1} \| \pi_{\theta_2} \big)\Big]$.
Recall that 
\begin{equation}
L_{\theta_{\text{old}}}(\theta) = \eta(\theta_{\text{old}}) + \sum_{s} \rho_{\theta_\text{old}}(s) \sum_{a} \pi_\theta(a|s) A_{\theta_\text{old}}(s,a).
\end{equation}
Since $\eta(\theta_{\text{old}})$ is constant, the condition in Eq.(\ref{eqn:conditionav}) simplifies to:
\begin{equation}
\begin{aligned}
&\underset{\theta}{\text{maximize}}~ \sum_{s} \rho_{\theta_\text{old}}(s) \sum_{a} \pi_\theta(a|s)  A_{\theta_\text{old}}(s,a),\\
&\text{subjected to} ~\overline{D}_{KL}^{\rho_{\theta_{\text{old}}}}(\theta_{\text{old}},\theta) \leq \delta.
\end{aligned}
\end{equation}

The following replacement is introduced:
\begin{equation}
\sum_s \rho(s) \rightarrow \mathbb{E}_{s \sim \rho},
\end{equation}
and using $Importance~Sampling$ \citep{Neal2001} done for a single state $s_n$ and described as
\begin{equation}
\begin{aligned}
&\sum_a \pi_\theta (a|s_n) A_{\theta_{\text{old}}}(s_n,a), \\
= &\sum_a q(a|s_n) \frac{\pi_\theta (a|s_n)}{q(a|s_n)} A_{\theta_{\text{old}}}(s_n,a),\\
= &~\mathbb{E}_{a \sim q} \Big [\frac{\pi_\theta (a|s_n)}{q(a|s_n)} A_{\theta_{\text{old}}}(s_n,a) \Big],
\end{aligned}
\end{equation}
where $q(a|s_n)$ is another simpler distribution, and $\frac{\pi_\theta (a|s_n)}{q(a|s_n)}$ is known as sampling weights. In the TRPO context, $q(a|s_n)$ is $\pi_{\theta_{\text{old}}}$.
Introducing these replacements, the condition thus becomes:

\begin{equation}
\begin{aligned}
&\underset{\theta}{\text{maximize}}~ \mathbb{E}_{a \sim \pi_{\theta_{\text{old}}}, s \sim \rho_{\theta _{\text{old}}}} \Big [\frac{\pi_\theta (a|s_n)}{\pi_{\theta_{\text{old}}}(a|s_n)} A_{\theta_{\text{old}}}(s_n,a) \Big],\\
&\text{subjected to} ~\mathbb{E}_{s \sim \rho_{\theta _{\text{old}}}} \big [D_{KL} \big( \pi_{\theta_{\text{old}}} \| \pi_{\theta}\big)\big] \leq \delta.
\end{aligned}
\end{equation}
By solving this condition and defining $\delta$ based on the problem at hand, the expected rewards at each iteration of solving will increase guaranteeing a better policy until optimization is reached.


\subsubsection{Divide and conquer reinforcement learning}

Finding the optimal policy for highly stochastic environments is a main challenge in reinforcement learning. High stochasticity is a synonym for wide diversity of initial states and goals and thus, it leads to a tedious learning process. Having TRPO presented, the work of \citep{ghosh2017divide} applied the algorithm on, for example, training a robotic arm to pick up a block and placing it in different positions. The idea behind their strategy is the following:
\begin{enumerate}
\item Slice the initial state-space into distinct slices.
\item Train each slice to find the corresponding optimal policy.
\item Merge the policies into a single optimal one that describes the entire space.
\end{enumerate}
This strategy of solving is named Divide-and-Conquer (DnC) reinforcement learning and is efficient for tasks with high diversity. Let's describe this algorithm.

Consider an MDP described as a tuple $\textbf{M} = (S, A, P, r, \rho)$. This MDP is modified to fit the "slicing" strategy i.e. the variables $\omega$ called contexts are introduced. The initial set is partitioned as $S = \bigcup_{i=1}^{n} S_i$ and each partition is associated to a context $\omega_i$ such that $\Omega = (\omega_i)^n_{i=1}$. Slicing is done using k-means clustering Appendix (\ref{k-mean}) for example. By that, $\rho$ becomes the joint probability distribution for the context and initial state set that is $\rho: \Omega \times S \rightarrow \mathbb{R}_{+}$.
Based on this slicing, the MDP extends into two:
\begin{enumerate}
\item Context-restricted MDP $\textbf{M}_\omega$: Given the context $\omega_i$, we find the policy $\pi_i(s,a) = \pi ((\omega_i, s), a)$, so $\textbf{M}_{\omega} = (S, A, P, r, \rho_{\omega}) $.
\item Augmented MDP \textbf{M'}: Each state is accompanied by a context ending with the tuple $(S \times \Omega, A, P, r, \rho)$, and the stochastic policy in this MDP is the family of context-restricted policies i.e. $\pi = (\pi_i)^n_{i=1}$.
\end{enumerate}
Finding the optimal policies in the context-restricted MDPs is finding the optimal policy in the augmented MDP $M_\omega$. Once the policy in the augmented MDP (whether the optimal policy or not) is found, the central policy $\pi_c$ in the original MDP (context-free MDP) can be found by defining $\pi_c(s,a) = \sum_{\omega \in \Omega} p(\omega|s)\pi_\omega(s,a)$. The main condition presented in this work is that the local policy in one context may generalize to other contexts; this accelerates the finding of the global policies that works for the original MDP which is context-independent.
Here's where TRPO kicks in. In order to find the optimal $\pi_c$ describing the original MDP, it is a must to find the augmented policy $\pi$ that maximizes the rewards more specifically, it maximizes 
\begin{equation}
\eta(\pi) - \alpha\mathbb{E}_\pi\big [D_{KL}(\pi\|\pi_c)\big],
\end{equation}
where $\alpha$ is the multiplier in order to integrate the condition back into the equation. Following the  TRPO regime for specifying the trust region, it is a condition that the policies in two respective contexts $\omega_i$ and $\omega_j$ should share as much information as possible, then:
\begin{equation}
\mathbb{E}_\pi\big [D_{KL}(\pi\|\pi_c)\big] \leq \sum_{i, j} \rho(\omega_i) \rho(\omega_j) \mathbb{E}_{\pi_i} \big [D_{KL}(\pi_i\|\pi_j)\big].
\label{eqn:centralKL}
\end{equation}

In the work of  \citep{ghosh2017divide}, instead of maximizing the surrogate function in Eq.(\ref{eqn:surrogate}), they consider it as a loss (multiplying by a minus sign) and aim to minimize it. Following this, the surrogate loss is:
\begin{equation}
\begin{aligned}
\mathcal{L}(\pi_1,&\dots , \pi_n) = - \sum_{i=1}^{n} \mathbb{E}_{\pi_{i, \text{old}}} \Big [\frac{\pi_i (a|s)}{\pi_{i,{\text{old}}}(a|s)} A_{i,{\text{old}}}(s,a) \Big]\\
&+ \alpha \Big(\sum_{i, j} \rho(\omega_i) \rho(\omega_j) \mathbb{E}_{\pi_i} \big [D_{KL}(\pi_i\|\pi_j)\big] \Big).
\end{aligned}
\end{equation}
Each policy is trained to take actions in its own context, but it is trained with the data of other contexts as well to ensure generalization. That being said, the surrogate loss for a single policy $\pi_i$ is 
\begin{equation}
\begin{aligned}
&\mathcal{L}(\pi_i) = - \mathbb{E}_{\pi_{i, \text{old}}} \Big [\frac{\pi_i (a|s)}{\pi_{i,{\text{old}}}(a|s)} A_{i,{\text{old}}}(s,a) \Big] +\\
&\alpha\rho(\omega_i) \sum_{ j}  \rho(\omega_j)
\Big(\mathbb{E}_{\pi_i} \big [D_{KL}(\pi_i\|\pi_j)\big] \\
&~~~~~~~~~~~~~~~~~~~~~~~+\mathbb{E}_{\pi_j} \big [D_{KL}(\pi_j\|\pi_i)\big]\Big).
\end{aligned}
\end{equation}
The steps of finding the optimal central policy is thus as follows. Within each context $\omega_i$, the local policy is enhanced using the surrogate loss with each iteration. After repeating this optimization procedure for several iterations, the central policy $\pi_c$ is found from the local policies by minimizing the KL-divergence of Eq.(\ref{eqn:centralKL}) which simplifies to
\begin{equation}
\begin{aligned}
\mathcal{L}_{\text{center}}(\pi_c) &= \mathbb{E}_\pi \big [D_{KL}(\pi\|\pi_c)\big]\\
&\propto \sum_i \rho(w_i) \mathbb{E}_{\pi_{i}} \big [ -\log \pi_c(s,a) \big].
\end{aligned}
\end{equation}
The TRPO algorithm accompanied with the constraints introduced here allowed to solved a highly stochastic MDP with diverse initial states and goals. The experimental work done in Divide-and-Conquer proves that this algorithm outperforms other RL algorithms.


\section{SCINET: A PHYSICS MACHINE }\label{ch4} 

We arrive now to introduce an example of a machine learning technique that empowers physics using neural networks. Approaches usually use experimental data and present them to neural networks in order to have it come up with the theory explaining the data. However, most techniques impose constraints on say, the space of initial states or the space of mathematical expressions. More specifically, these techniques incorporate our physical intuition to the neural network, thus mainly they're testing the network's efficiency and learnability rather than its ability to output theories from scratch. In the work of \citep{iten2018discovering}, this problem is tackled by constructing a neural network named $SciNet$ on which no constraints or any previous knowledge are applied. $SciNet$ must as well output the parameters that describe the physical setting wholly and sufficiently.
The idea presented in this work is as follows:
\begin{enumerate}
\item Supplying $SciNet$ with experimental data,
\item $SciNet$ finds a simple representation of the data,
\item then a question will be asked for $SciNet$ to answer.
\end{enumerate}

$SciNet$ must be able to answer the question using only the representation it gave without going back to the input data. These steps are approached using two models:
\begin{itemize}
\item Encoder: The encoder structure is made of one or more neural networks. It takes the observations $\mathcal{O}$ (experimental data) and encodes them into representations $\mathcal{R}$ named latent representations in machine learning context. The mapping is thus $\mathcal{E}: \mathcal{O} \rightarrow \mathcal{R}$.
\item Decoder: The decoder structure is also a neural network. It takes as inputs the latent (hidden) representations $\mathcal{R}$ produced by the encoder as well as the question $\mathcal{Q}$ to be answered. It outputs the answer to the question. The mapping is thus $\mathcal{D}: \mathcal{R} \times \mathcal{Q} \rightarrow \mathcal{A}$.
\end{itemize}

\begin{figure}
    \centering
    \includegraphics[totalheight=4.5cm]{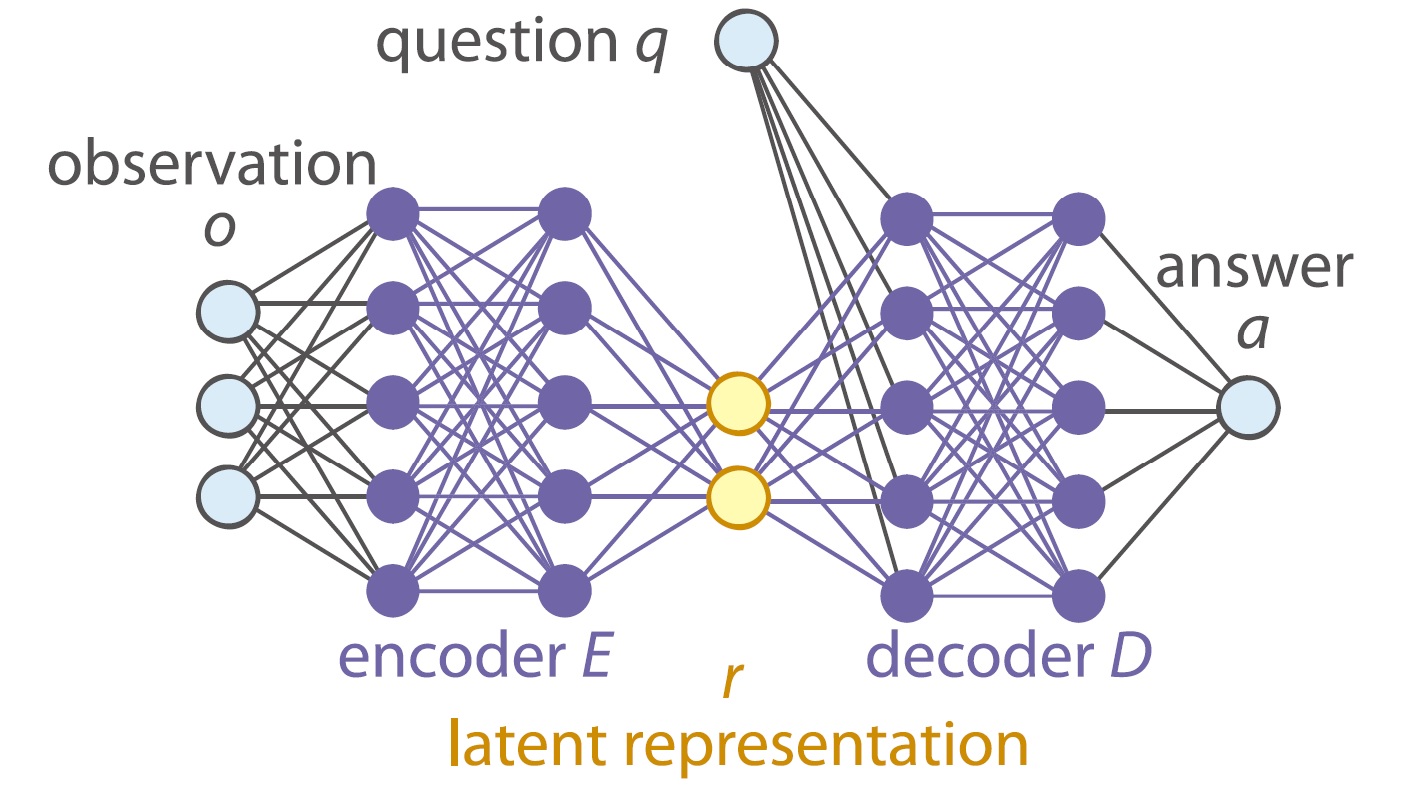}
    \caption{Illustration of SciNet \citep{iten2018discovering}}
    \label{fig:scinet}
\end{figure}

Fig.(\ref{fig:scinet}) illustrates the encoder and decoder networks.

$SciNet$'s encoder and decoder are trained with a chosen training set of observations and questions, and then they are tested with the chosen test set to predict the accuracy. We must note that since we don't previously know or impose the number of latent neurons (those specific for the latent representation), the accuracy of prediction may be low due to insufficient latent neurons. In that sense and during the training phase, the number of latent neurons may be reset to fit the representations.

As a simple example, suppose that you feed $SciNet$ with observations of the variation of the electric potential $U$ as a function of current $I$ governed by Ohm's law. But, $SciNet$ has no idea what Ohm's law is; it is only seeing the introduced observations. The encoder will find a representation for these observations which is the parameter $R$ standing for resistance and store it in a latent neuron. Being supplied with the representation and a question as what will the potential be for a given current, the decoder predicts the right answer.

Let's introduce some of the examples presented in the paper that demonstrates $SciNet$'s efficiency and accuracy in predicting representations and answers from scratch without any constraints given.


\subsection{Experiment One: Damped Pendulum}
Presented as a simple classical example of how $SciNet$ works, the double pendulum is described by the following differential equation
\begin{equation}
-\kappa  x -b\dot{x} = m\ddot{x},
\end{equation}
where $\kappa$ is the spring constant which governs the frequency of oscillation $\omega = \sqrt{\frac{k}{m}}\sqrt{1- \frac{b^2}{4m\kappa}}$ and $b$ is the damping factor. The solution of the damped system is 
\begin{equation}
x(t) = A_0 e^{-\frac{b}{2m}t} \cos(\omega t +\delta_0).
\end{equation}

$SciNet$ is implemented as a network with three latent neurons, and it is fed with time-series observations about the position of the pendulum. The amplitude $A_0$, mass $m=1$, and phase $\delta_0=0$ are fixed for all training sets only the spring constant $\kappa$ and damping factor $b$ vary between $[5, 10] kg/s^2$ and $[0.5, 1] kg/s$ respectively.

The encoder outputs the parameters $\kappa$ and $b$ and stores them in two latent neurons without using the third neuron. Upon providing time $t_{\text{pred}}$ as a question, $SciNet$ predicts through the decoder network the position of the pendulum at that time with excellent accuracy.
Therefore, $SciNet$ was able to draw out the physical parameters and store them as well as predict future positions accurately. This implies that the parameters extracted where sufficient to describe the whole system as well as make future predictions.


\subsection{Experiment Two: Qubits}
After presenting a classical example, $SciNet$ is tested with quantum examples, specifically with qubits. Before heading to explain the problem at hand, we define a couple of terminologies used:
\begin{enumerate}
\item \underline{Qubit}: A qubit, the quantum analog of a classical bit, is a two-dimensional system that can exist in a superposition of two states. It forms the fundamental unit in quantum computing.
\item \underline{Quantum tomography}: It is a method of reconstructing the quantum state from a series of measurements \citep{paris2004quantum}. A typical approach is to prepare copies of the quantum state and perform several measurements on the copies. Each of these measurements allows us to have part of the information stored in the state. If the set of measurements is informationally complete thus allowing reconstruction of the quantum state fully, they are said to be tomographically complete. Otherwise, they are tomographically incomplete. 
\item \underline{Binary projective measurements}: Measurements to yield the state of the qubit: $ |0\rangle$ or $|1\rangle$ for a single qubit for example.
\end{enumerate}

Given a set of measurements , $SciNet$ is required to represent the state of the quantum system as well as make accurate predictions without having any previous quantum knowledge. We suppose in this example that the two considered states, to be represented, are a 1-qubit state and 2-qubit state. The number of real parameters for a single qubit is two, and that for double qubits is six. Here's how we got those numbers:
\begin{itemize}
\item The dimension for the complex vector space is $2^n$ for $n-$qubits, so for a single qubit we have two states: $|0\rangle$ and $|1\rangle$; for double-qubits we have four states: $|00\rangle$, $|01\rangle$, $|10\rangle$ and $|11\rangle$.
\item Counting the number of real parameters, then we'll have $2\times 2^n$.
\item Two constraints are present. The first is the normalization condition $\langle\psi|\psi\rangle = 1$, and the second is that the global phase factor doesn't hold any information meaning that having $\psi ' = e^{i\phi}\psi$ will not affect the inner product. These constraints will lessen the number of parameters by two.
\end{itemize}

Therefore, we should expect from the encoder to use two latent parameters for 1-qubit and six latent parameters for 2-qubits. From the set of all binary projective measurements $\mathcal{M}$, a random subset $\mathcal{M}_1 = \{\alpha_1, \alpha_2,\dots, \alpha_{n_1}\}$ ($n_1$ = 10 for single-qubit, and 30 for 2-qubits) is chosen and projected on $\psi$ where $\psi$ is the quantum state to be represented. The probabilities generated $p(\alpha_i, \psi)$'s are the probabilities of measuring zero. After repeating the measurements several times, the resulting probabilities are fed to the network as observations. Given these observation, $SciNet$ determines the minimal number of parameters sufficient to describe the quantum state.

Choosing another random set of binary projective measurements $\mathcal{M}_2 = \{\beta_1, \beta_2,\dots, \beta_{n_2}\}$ ($n_2$ = 10 for single-qubit, and 30 for 2-qubits), we project these measurements on another measurement $\omega$ to generate the set of probabilities $p(\beta_i, \omega)$. These probabilities are fed to $SciNet$ as a question. Note that $\mathcal{M}_2$ is taken to be tomographically complete, that is $\omega$ is described fully, while $\mathcal{M}_1$ may be complete or incomplete. $SciNet$ is required to predict the probability $p(\omega, \psi) = |\langle\omega, \psi\rangle|^2$ of applying $\omega$ to measure zero.

For the 1-qubit system and given the sufficient number of latent neurons (two neurons), $SciNet$ predicted the probability with very high accuracy given that $\mathcal{M}_1$ is tomographically complete. When $\mathcal{M}_1$ was tomographically incomplete, the error percentage was somewhat high. Tomographically incomplete set is recognized since the accuracy wasn't high, and the accuracy percentage gave an idea of the amount of information provided by the set.

For the 2-qubit system, the percentage of error was very low given a sufficient number of neurons (minimum of six) and a tomographically complete set. For a tomographically incomplete set, the percentage of error was high no matter how many latent neurons are added.


\subsection{Experiment Three: Heliocentric Model}
The previous examples were time-independent models. Dealing with time-dependent ones as done here requires a slight modification with the structure of $SciNet$. Given the observations, the encoder outputs the representations for the initial time $r(t_0)$. These representations evolve in time to become $r(t_1)$ and so on. At each time step, the decoder network outputs an answer.

The idea behind this experiment is that given the angles of Mars and Sun as seen from Earth , $SciNet$ will predict these angles at each time step. When we say angles as seen from Earth, it is as if we're considering the geocentric model (Earth in the center of the Solar system). However, given a set of angles of Mars $\theta_M$ and the Sun $\theta_S$ as seen from Earth, $SciNet$ stores the angles of Earth $\phi_E$ and Mars $\phi_M$ as seen from the Sun. $SciNet$ constructs the Heliocentric model. Two neurons are thus activated to save $\phi_E$ and $\phi_M$. The decoder network receives these representations at each time step $\theta_i$ and outputs the $\theta_M(t_i)$ and $\theta_S(t_i)$ as required with very low error percentage (less than 0.4\%).

$SciNet$ as a neural network succeeded to represent physical systems in the sufficient minimal number of physical parameters given no background about the system. It is a concrete example of how physics can benefit from AI and neural networks to achieve desired results. 


\section{THE AI PHYSICIST}\label{ch5}

It is worth mentioning, first, a distinguished particularity concerning the ML techniques implemented in the work of \citep{iten2018discovering}. The built neural networks are fed with data sets that describe a specific physical setting. Their role, as thoroughly detailed in section (\ref{ch4}) , is to come up with a physical theory describing the input data. Researchers from MIT also adopted the same theme to improve our physical intuition using ML techniques. However, they proposed a different question: Is it possible to produce an AI system that infers theories describing different aspects of the world, all at once? To address this question, Tegmark $et~al.$ in \citep{wu2018toward} introduce an ML algorithm that mimics the human scientist's way of thinking. They call it the "AI Physicist". The ``AI Physicist" agent successfully learns theories and uses them to infer future domain-specific predictions. The ``AI Physicist" tackles a physical problem with various complexities using four consecutive strategies, three of which are: divide-and-conquer, Occam's razor, and unification. An additional concern for an intelligent agent is the ability to learn faster, i.e. reaching the desired accuracy through high learning speeds. However, a main problem that faces agents is forgetting the already learned tasks when being trained for a new one. The phenomenon of forgetting previous tasks is called catastrophic forgetting. The ``AI Physicist" made the attempt to overcome it through its fourth and final strategy known as Lifelong Learning. This approach is implemented successfully using the Theory Hub which mimics a hard computer's memory.


\subsection{Architecture}
The ``AI Physicist" processes the given data as following:

\begin{enumerate}
\item The theory hub proposes theories from the space of previously saved theories. These theories describe parts of the data points, and new theories are randomly initialized to regard the rest.
\item The divide-and-conquer algorithm trains proposed and newly discovered theories to best fit the data. They are trained first all together to minimize a generalized mean loss then separately to fine-tune each one specifically to the domain it describes. 
\item Well-defined theories are then added to the theory hub.
\item The Occam's razor Algorithm organizes the fine-tuned theories to transform them into simpler, symbolic expressions.
\item The Unification algorithm joins symbolic theories into master ones. 
\item The master and symbolic theories are then added to the theory hub and proposed when a new environment is encountered.
\end{enumerate}

\begin{figure}
     \centering
     \includegraphics[totalheight=5.7cm]{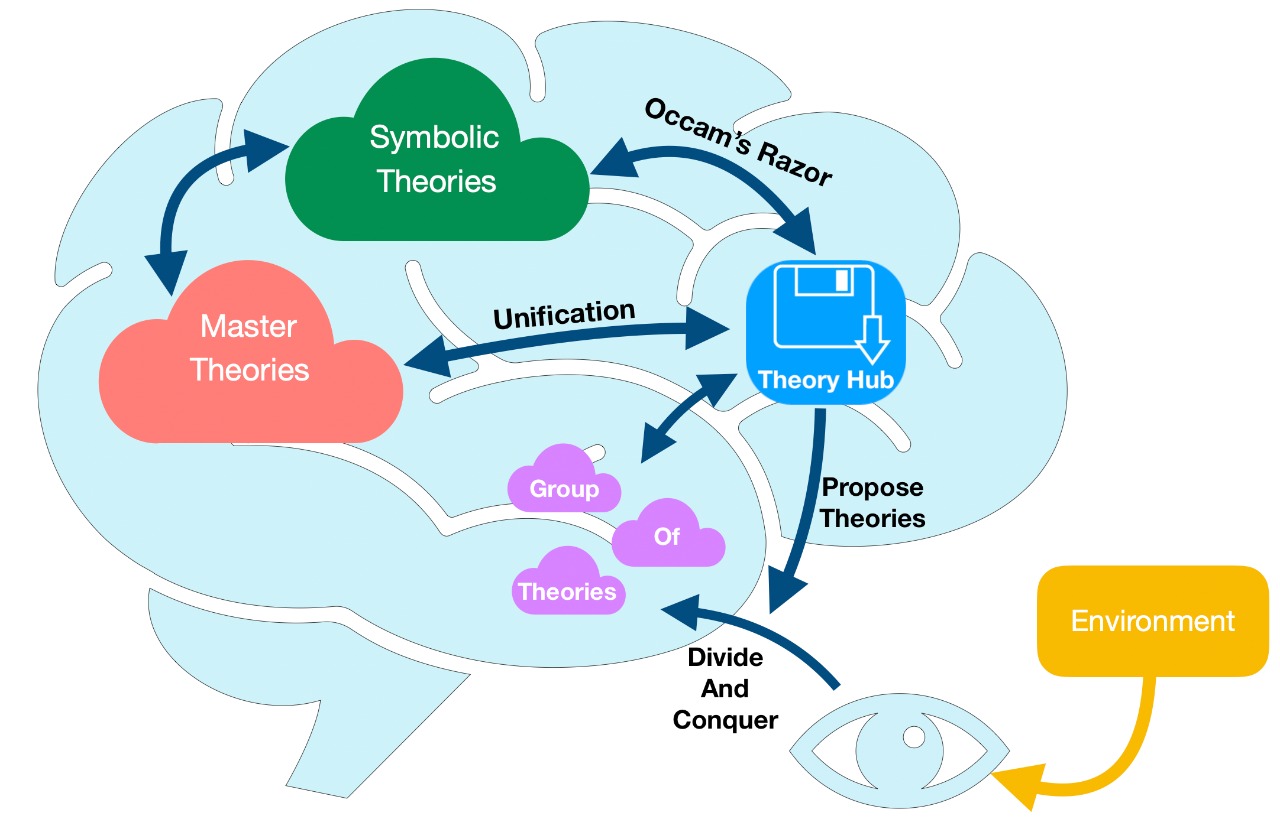}
     \caption{The ``AI Physicist''}
     \label{fig:MIT}
 \end{figure}
 
All these steps can be figured out in the illustration found in Fig.(\ref{fig:MIT}).

Before going into the details, it is necessary to highlight on some remarks that help understand the strategies. First, the data $D$ given to the ``AI Physicist" is characterized as a time series such that $D=\{(\mathbf{x}_t,\mathbf{y}_t)\}=\{(\mathbf{x}_{t-T},\dots,\mathbf{y}_{t-1},\mathbf{y}_t)\}$ consisting of $T$ vectors. A theory $\mathcal{T}$ is defined as a 2-tuple $(\textbf{f} ,c)$. $\textbf{f}$ is the prediction function that maps a data point at a certain time step to another one at a further time step and $c$ is the domain sub-classifier which classifies each data point into its related domain. $\textbf{f}$ and $c$ are both implemented as neural networks with learnable parameters, such that the neural network of $\textbf{f}$ consists of two hidden layers with linear activation and that of the sub-classifier has two hidden layers with leakyReLU activation and an output layer with linear activation.


\subsection{The Learning Algorithm}


\subsubsection{Proposing theories}

\begin{algorithm}[H]
\caption{\textbf{Theory Proposing from Hub}}
\begin{algorithmic}[1]
\Require  \textbf{Hub}: theory hub
\Require Dataset $D=\{(\mathbf{x_t},\mathbf{y_t})\}$
\Require $M_0$: number of theories to propose from the hub
\State  $\{(\mathbf{f}_i, c_i)\} \leftarrow$ \textbf{Hub}.retrieve-all-theories()
\State  $D_{\mathsf{best}}^{(i)}\leftarrow \{(\mathbf{x_t},\mathbf{y_t})|\arg\min_j \ell_{DL,\epsilon}[\mathbf{f}_j(\mathbf{x}_t),\mathbf{y}_t]=i\}, \forall i$
\State  $\mathcal{T}_{M_0}\leftarrow \big\{(\mathbf{f}_i, c_i)| D_{\mathsf{best}}^{(i)}$ ranks among $M_0$ largest sets in $\{D_{\mathsf{best}}^{(i)}\} \big\}$
\newline
\State \Return $\mathcal{T}_{M_0}$
\end{algorithmic}
\end{algorithm}

When the ''AI Physicist" is given a new data describing an environment it didn't encounter previously, it first proposes theories from the theory hub that describes part of the data. This is a key evidence that past theories are not forgotten and are reused at a later time. This is a part of the lifelong learning strategy. From a total of $M$ theories describing the set, $M_o \leq M$ theories must be proposed from the theory hub. $M$ and $M_o$ are previously specified. Initially, when the data set $D=\{(\mathbf{x}_t,\mathbf{y}_t)\}$ is supplied to the agent, the theory hub examines all theories first (step 1). For each data point $(x_t, y_t)$, the agent saves the index $i$ of the theory that best describes this point through minimizing the corresponding loss function (step 2). Any other data point that accounts to the same theory is added to the previous one and all corresponding points are put together into a data subset noted as $D^{(i)}$. In this vein, several subsets \{$D^{(i)}$\} are constituted and for each one the number of data points $n_i$ is counted. Among all of them, the $M_o$ sets with the largest $n_i$, and therefore their corresponding theories, are then proposed.


\subsubsection{Divide-and-conquer algorithm}
The environment introduced to the agent is described by several theories. As previously mentioned, the theory consists of its domain sub-classifier ${c}$ and the function $\bf{f}$. The idea behind the algorithm is to: divide the data into domains using the sub-classifier and then train each function in the domain it conquers. The function $\bf{f}$ maps $\textbf{x}_{t} \rightarrow \textbf{y}_{t}$, and it is parameterized by a parameter vector $\theta$. Recall that $\bf{f}$ and ${c}$ are implemented as neural networks and the parameters of neural networks are the weights and biases. The parameters are learned and adjusted using gradient descent in order to minimize the following loss
\begin{equation}
\mathcal{L} = \sum_{t} \ell [\textbf{f}(\textbf{x}_t),\textbf{y}_t)],
\end{equation}
where $\ell$ is the distance between the output resulting from the prediction function $\textbf{f}(\textbf{x}_t)$ and the output $\textbf{y}_t$.

However, since the environment encountered is a mixed one where several theories are found, each theory should compete with the other theories in specializing in its own domain with the generalized mean-loss being
\begin{equation}\label{Eq5.2}
\mathcal{L}_{\gamma} = \sum_{t}\Big ( \frac{1}{M} \sum_{i=1}^{M} \ell [\mathbf{f}_i(\textbf{x}_t,\textbf{y}_t)]^{\gamma} \Big )^{\frac{1}{\gamma}}.
\end{equation}
This loss is dependent on $\gamma$. For a negative value of $\gamma$, the function that best decreases the loss function $\ell$ will dominate $\mathcal{L}_\gamma$. In other words, the function $\mathbf{f}_i$ that best predicts the output will be dominant. As $\gamma$ becomes more negative, $\mathcal{L}_{\gamma}$ will tend to be equal to the minimum loss resulting from the well-fitted function $\mathbf{f}_i$, i.e
\begin{equation}
\gamma \rightarrow -\infty  \Rightarrow\mathcal{L}_{\gamma} \rightarrow  \min \ell [\mathbf{f}_i (x_t, y_t)].
\end{equation}
As each $\mathbf{f}_i$ best fits its domain, the generalized mean loss is minimized. For further technical details, check Appendix (F) in \citep{wu2018toward}.

It is empirically found that $\gamma = -1$ suits the procedure. It works well in the process of specializing the functions and in giving the gradient for improving other theories during gradient descent as will appear in the algorithm. The loss $\mathcal{L}_{-1}$ is termed harmonic mean loss (harmonic mean is the reciprocal of the arithmetic mean of reciprocals).

Finding the function that best fits the data in its domain is the process of minimizing the error at two levels. The first is minimizing the harmonic mean loss $\mathcal{L}_{-1}$ that includes all functions at once. The second is minimizing the loss $\ell [\mathbf{f}_i(\mathbf{x}_t,\mathbf{y}_t)]$ specific to each function in its domain as a way of fine-tuning each function. We choose the $\ell$ to be the description length (DL) loss function :
\begin{equation}
\ell(u_t) = DL(u_t)=  \log_2 \Big(1+(\frac{u_t}{\epsilon})^2\Big),
\label{eq:loss}
\end{equation}
where $u_t = | \mathbf{f}(\mathbf{x}_t)-\mathbf{y}_t|$. The description length is the number of bits required for storing, and we aim to lower this number as much as possible. By lowering the loss $\ell$, we lower the DL. Regarding the sub-classifiers ${c}$, the loss that we aim to minimize is the categorical cross-entropy loss. This loss combines the softmax activation with the cross entropy loss. The softmax activation is usually used for classification in such a way it outputs a probability distribution among all predicted classes. The cross entropy then checks the effectiveness of this classification. This means that if a required class acquires a low probability, high cross entropy loss will result indicating bad classification.

The algorithm developed for divide-and-conquer is named unsupervised differentiable divide-and-conquer (DDAC) algorithm. For the dataset $D= \{ (\mathbf{x}_t, \mathbf{y}_t) \}$, the number $M$ is the number of initial theories to be trained out of which $M_0$ theories are proposed by the theory hub (authors in \citep{wu2018toward} used $M=4$ and $M_0=2$). The initial precision error $\epsilon_0$ is set to be quite large (10 is used) so that the loss $\ell(u_t) =  \log_2 \Big(1+(\frac{u_t}{\epsilon })^2\Big ) \approx u_t^2$ becomes the quadratic mean-squared error (MSE). After each iteration, the error is set is to be the median prediction error. Dealing with the loss function in equation Eq.(\ref{eq:loss}) is quite tricky and tedious that is why having it approximated to be MSE is better.

The algorithm starts by randomly initializing $M-M_0$ theories that are not proposed by the theory hub, thus having $M$ theories: $\bf\mathcal{T} = \{\mathcal{T}_1,..., \mathcal{T}_M\}$ and the functions $\bf{f}$ are parameterized by a vector $\theta$: $\bf{f_\theta}$. The sub-classifiers ${c}$ are parameterized by the vector $\phi$: ${c_\phi}$ where $\theta$ and $\phi$ are learnable parameters.
The first stage is to train the theories using the harmonic loss $\mathcal{L}_{-1}$ (steps 2-3). The harmonic loss is passed as an attribute to the subroutine IterativeTrain which forms the central part of this algorithm.

Within the steps of subroutine IterativeTrain (steps s1-s10), the gradient of the harmonic loss with respect to the parameter $\theta$ is computed and used in stochastic gradient descent or Adam to update it with a learning rate $\beta_f$ (the step size of updates; $5 \times 10^{-3}$ is used). For a $\mathbf(\textbf{x}_t,\textbf{y}_t)$ pair, the index $i$ of the function $\mathbf{f_i}$ that minimizes the loss is saved in the parameter $b_t$.

\begin{algorithm}[H]

\caption{\textbf{Differentiable Divide-and-Conquer with Harmonic Loss}}
\begin{algorithmic}[1] 
\label{ddac}
\Require Dataset $D=\{(\mathbf{x_t},\mathbf{y_t})\}$
\Require M: number of initial total theories for training
\Require $\mathcal{T}_{M_0}=\{(\mathbf{f_i},c_i)\}$, $i= 1,\dots,M_0$, $0\leq M_0 \leq M$: theories proposed from theory hub
\Require $K$: number of gradient iterations
\Require $\beta_{\mathbf{f}},\beta_{\mathbf{c}}$: learning rates
\Require $\epsilon_0$: initial precision floor
\newline
\State Randomly initialize $M-M_0$ theories $\mathcal{T}_i$, $i=M_0+1,\dots, M$. Denote $\mathcal{T}_i=(\mathcal{T}_1,\dots, \mathcal{T}_M)$, $\mathbf{f}_{\theta} =(\mathbf{f}_{1}, \dots ,\mathbf{f}_{M})$, $\mathbf{c}_{\phi}=(c_1,\dots, c_M)$ with learnable parameters $\theta$ and $\phi$.

\ 
\Statex // \textit{Harmonic training with DL loss:}
\State $\epsilon  \leftarrow \epsilon_0 $
  \For{\textit{k} \textbf{in} $\{1,2,3,4,5\}$}
  \State $\mathcal{T} \leftarrow $ \textbf{IterativeTrain}($\mathcal{T},D,\ell_{DL,\epsilon},\mathcal{L}_{-1}$), 
  \newline where $\mathcal{L}_{-1}\equiv \sum_t\left({1\over M}\sum_{i=1}^M\ell [\mathbf{f}_i(\mathbf{x}_t),\mathbf{y}_t]^{-1} \right)^{-1}$ (Eq.\ref{Eq5.2})
   \State $\epsilon \leftarrow$ set\_epsilon($\mathcal{T},D$) //\textit{median prediction error}
  \EndFor

\
\Statex//\textit{Fine-tune each theory and its domain:}
\For{\textit{k} \textbf{in} $\{1,2\}$}
\State $\mathcal{T} \leftarrow $ \textbf{IterativeTrain}($\mathcal{T},D,\ell_{DL,\epsilon},\mathcal{L}_{\mathsf{dom}}$), 
   where \newline $\mathcal{L}_{\mathsf{dom}}\equiv \sum_t\ell [\mathbf{f}_{i_t}(\mathbf{x}_t),\mathbf{y}_t] $ with $i_t=\arg \max _i \mathbf{c}_i(\mathbf{x}_t ) $
\State$\epsilon \leftarrow$ set\_epsilon($\mathcal{T},D$) //\textit{median prediction error}

\EndFor

\State  \Return $\mathcal{T}$
\newline
\end{algorithmic}

\textbf{subroutine IterativeTrain}($\mathcal{T},D,\ell,\mathcal{L}$)
\begin{algorithmic}[1] 
\algrenewcommand{\alglinenumber}[1]{\bfseries s{#1}:}
\renewcommand{\Statex}{\item[\hphantom{\bfseries Step \arabic{ALG@line}.}]}

\For{\textit{k} \textbf{in} $\{1,\dots, K\}$}
 \Statex // \textit{Gradient descent on} $\mathbf{f}_{\theta}$ \textit{with loss} $\mathcal{L}$:
\State $\mathbf{g}_{\mathbf{f}}\leftarrow \nabla_{\theta}\mathcal{L}[\mathcal{T},D,\ell]$
\State Update $\theta$ using gradients $\mathbf{g}_{\mathbf{f}}$ (e.g: Adam \citep{kingma2015adam})
\newline
\Statex\textit{// Gradient descent on $\mathbf{c}_\phi$ with the best performing theory index as target:}
\State $b_t \leftarrow  \arg \max _i  \left\{ \ell [\mathbf{f}_{i_t}(\mathbf{x}_t),\mathbf{y}_t] \right\}$, $\forall t$

\State $\mathbf{g}_{\mathbf{c}}$ $\leftarrow \nabla_{\phi}\sum_{(\mathbf{x}_t,\cdot)\in D}$CrossEntropy[softmax$(\mathbf{c}_{\phi}(\mathbf{x}_t)),b_t$]
\State Update $\phi$ using gradients $\mathbf{g}_{\mathbf{c}}$ (e.g: Adam \citep{kingma2015adam} or SGD \citep{nielsen2015neural})
\EndFor
\State $\mathcal{T}\leftarrow$ AddTheories$(\mathcal{T},D,\ell,\mathcal{L})$ //Optional 
\State $\mathcal{T}\leftarrow$ DeleteTheories$(\mathcal{T},D,\ell)$ //Optional 
\State  \Return $\mathcal{T}$ 
\end{algorithmic}
\end{algorithm}
 In this way, we're classifying each data set by the best-fitted function. After training the functions of the theories to minimize the loss, the sub-classifiers are trained. In step s6 of the IteravtiveTrain, the loss that we need to minimize is the cross-entropy, and to do that, we need to transform the outputs of the sub-classifier ${c_\phi}$ into probabilities. This is done using the softmax function. After applying the softmax function to the outputs of ${c_\phi(x_t)}$, they are implemented in the cross-entropy loss function along with $b_t$. The gradient of this loss is computed and used in SGD or Adam to update the parameters $\phi$ of the sub-classifier with a learning rate $\beta_c$ ($10^{-3}$ is used). These steps of training the functions and sub-classifiers using the harmonic loss $\mathcal{L}_{-1}$ and the cross-entropy are repeated iteratively (authors used K=10000 iterations).

 Step (s8) in the IterativeTrain checks the theories that describe a fair set of data (the authors used a threshold of 30\%), and if a certain fraction of this set (threshold 5\%) is found to have a MSE greater than a certain number ($2\times10^{-6}$ is used), then another theory $\mathcal{T}_{M+1}$ is initialized and passes through steps s1- s6 of the IterativeTrain. The loss of the the $\mathcal{T}_{M+1}$ is computed. If the loss is greater than what it was before adding the theory, the new theory is rejected. Otherwise, it is accepted and trained. Step (s9) in the IterativeTrain checks the theories once again. If a theory describes a small fraction of the dataset (0.5\% is used), it is deleted.

The second stage of fine-tuning each function in its domain follows the same steps as the first stage but using the loss $\ell$. Each stage is also done iteratively and after each iteration, the precision error is computed (lower value).


\subsubsection{Adding theories}
\begin{algorithm}[H]
\caption{\textbf{Adding Theories to Hub}}
\begin{algorithmic}[1]
\label{adding theories}
\Require  \textbf{Hub}: theory hub
\Require  $\mathcal{T}=\{(\mathbf{f_i},c_i)\}$: Trained theories Alg.2
\Require  Dataset $D=\{(\mathbf{x_t},\mathbf{y_t})\}$
\Require  $\eta$: DL threshold for adding theories to hub
\State $D^{(i)}\leftarrow \{(\mathbf{x_t},\mathbf{y_t})|\arg\max_j \{c_j(\mathbf{x}_t) \} =i\},\forall i $
\State dl$^{(i)} \leftarrow{1\over |D^{(i)}|} \sum_{(\mathbf{x}_t,\mathbf{y}_t)\in D^{(i)}} \ell_{DL,\epsilon} [\mathbf{f}_i(\mathbf{x}_t),\mathbf{y}_t],\forall i$

\For {\textit{i} \textbf{in} $\{1,2,\dots \mathcal{|T|}\}$ }
\If{dl$^{(i)}<\eta$} 
\State \textbf{Hub}.addIndividualTheory$((\mathbf{f}_i, c_i),D^{(i)})$
\EndIf
\EndFor
\end{algorithmic}
\end{algorithm}
To add trained theories to the theory hub after the divide-and-conquer algorithm, the description length dl of each theory in its domain is first calculated. Whenever this description length does not exceed a certain threshold $\eta$, the theory is directly added to the theory hub. Not only that, but also the corresponding data that are fit by each theory is also added for the sake of knowing how the theory had been previously trained. This is also beneficial when applying the following algorithm Occam's Razor as will be shown in the next section.


\subsubsection{Occam's razor}
\paragraph{Overview}

Making inferences based on a limited data set faces a serious problem known as model selection. This problem arises whenever there is a need to choose one hypothesis, among other competing hypotheses, that best explains the given data. Occam's Razor plays an important role to solve this conflict. Formally stated, Occam's Razor states that among competing hypotheses, the one with the least number of assumptions is the one that best fits the data. This means that the data under study is usually modeled with the simplest explanation.

Occam's Razor is used in a wide range of domains to rule out any unnecessary information or elements. Restricting the discussion to the scientific domain, physicists frequently use Occam's razor. Here are two examples illustrating Occam's Razor \citep{or} :
\begin{enumerate}
    \item Geocentric vs. Heliocentric Models of the Solar System: There exist an old debate in astronomy about modelling the solar system. Two theories competed for that, the first known as the geocentric model which states that the Sun and the planets are in orbit around the stationary Earth and the other known as heliocentric model in which the Sun is the center of orbiting of all planets.

  Occam's razor turned off the discussion when applying it. Copernicus, who developed the heliocentric model many years after developing the geocentric one won the competition because his theory was much simpler. The latter included more sophisticated assumptions in addition to the presence of several unexplained mysteries.
  
  \item Einstein vs. Lorentz: In the 20th century, there appeared an attempt to explain the space-time continuum. Two physicists at that time, Albert Einstein and Hendrik Lorentz addressed this issue and made up all the mathematical explanations but with different calculations.

Lorentz based his calculations on the assumption that there exists a motionless medium in space known as ether while Einstein made his explanations without any reference to the ether. Because no experimental evidence was made for the presence of such a medium and accepting Occam's razor, Einstein won the competition.
\end{enumerate}

\paragraph{Minimum Description Length Formalism}
After having all theories learnt, they are treated with Occam's Razor strategy. One form of this strategy includes a mathematical formalism known as the minimum-description-length (MDL) formalism.

The main originator of MDL formalism was Quoting Rissanen in 1989 \citep{grunwald2005advances}. Given a data set, Rissanen defines the number of bits to describe the data as the description length (DL). The main insight for this formalism is centered around two ideas, the first is detecting any regularity (e.g. a pattern) in the data and employing it to compress these data. These regularities shape the properties of the data. In that way, meaningful information is revealed. The second central idea states that learning from the data occurs whenever a regularity is discovered. This means that the more we compress the data, the more we learn from it. Rissanen interprets the model that describes the regularities as the program that produces the data as an output. Of course, the model is identified also with its corresponding code. Hence, the description length is now defined as the number of bits of that program including the compressed data bits.

\paragraph{Occam's Razor Algorithm}
In the context of the "AI Physicist", the main goal of Occam's Razor is to minimize the description length of the prediction functions $\mathbf{f}_i$'s obtained by the DDAC algorithm using the MDL formalism. The domain sub-classifier is unique for each domain, thus it's not practical to be concerned in minimizing its DL. On the other hand, the importance of the prediction functions is that it can be reused in other domains or to solve new incoming data. First, the algorithm calculates the description length of theories in hand then tries to minimize it.

For a dataset $D=\{(\textbf{x}_t,\textbf{y}_t)\}$, the trained theories are $\mathcal{T}=\{(f_i,c_i)\}$. The DL of such theories is the DL of each theory plus the DL of its error:
\begin{equation}
DL(\mathcal{T},D)=DL(\mathcal{T})+\sum_t DL(u_t),
\label{1}
\end{equation}
with $u_t= |\hat{y}_t-y_t|$. The second term in Eq.(\ref{1}) has been minimized by the DDAC algorithm. Occam's Razor algorithm thus focuses on minimizing $DL(\mathcal{T})$ which can be decomposed as $DL(\mathcal{T})=DL(f_{\theta})+DL(c_{\phi})$ where $f_{\theta}=(f_1,...f_M)$ and $c_{\phi}=(c_1,...c_M)$. The $DL(f_\theta)$ can be defined as the description length of the parameters $\theta$ of $f$:
\begin{equation}
DL(f_\theta)=\sum_j(\theta _j). 
\end{equation}

To minimize the DL of the parameters of $f_\theta$, transformations are applied such as collapseLayers, localSnap, integerSnap and others. These transformations will keep rolling as long as they minimize the total description length $DL(\mathcal{T}, D)$. These transformations will be detailed and explained thoroughly.

The algorithm starts by decomposing the dataset $D$ into subsets $D^{(i)}$ (step 2) each in its corresponding domain. Then it applies the first transformation collapseLayers (step 3) which finds all successive layers of a neural network having linear activation and combines them. As mentioned previously, the prediction function predicts the positions at the current time step from previous ones. The localSnap transformation (step 4) interferes to take into account only inputs closer to the current time step to be predicted.  The intergerSnap transformation (step 5) transforms the parameters in $f_i$ into an integer iff this minimizes the total DL. For example, given a parameter $p=1.99992$, the transformation will replace it by $p=2$ and then calculate the new description length. If this snap minimizes the DL, it takes it otherwise leaves it.  Another example with $p=1.6666633$, the algorithm will try $p=2$ and calculate the DL. If it finds that the DL has increased, it will leave it and try another transformation which is the rationalSnap that can approximate $p=1.5={3\over2}$. To calculate the description length of an integer $m$ we use:
\begin{equation}
DL(m)=\log_2(1+|m|).
\end{equation}

\begin{algorithm}[H]
\caption{\textbf{Occam's Razor with MDL}}
\begin{algorithmic}[1] \label{occam's razor}
\Require Dataset $D=\{(\mathbf{x_t},\mathbf{y_t})\}$
\Require $\mathcal{T}_{M_0}=\{(\mathbf{f_i},c_i)\}$, $i= 1,\dots,M_0$, theories trained after Alg.2
\Require $\epsilon$: precision floor for $\ell _{DL,\epsilon}$
\For{\textit{i} \textbf{in} $\{1,\dots, M\}$}
\State $D^{(i)} \leftarrow \left\{(\mathbf{x}_t,\mathbf{y}_t)| \arg \max _j \{\mathbf{c}_j(\mathbf{x}_t)\}=i   \right\}$, 
\State $\mathbf{f}_i\leftarrow$ \textbf{MinimizeDL}(collapseLayer, $\mathbf{f}_i, D^{(i)}, \epsilon$)
\State $\mathbf{f}_i\leftarrow$ \textbf{MinimizeDL}(localSnap, $\mathbf{f}_i, D^{(i)}, \epsilon$)
\State $\mathbf{f}_i\leftarrow$ \textbf{MinimizeDL}(integerSnap, $\mathbf{f}_i, D^{(i)}, \epsilon$)
\State $\mathbf{f}_i\leftarrow$ \textbf{MinimizeDL}(rationalSnap, $\mathbf{f}_i, D^{(i)}, \epsilon$)
\State $\mathbf{f}_i\leftarrow$ \textbf{MinimizeDL}(toSymbolic, $\mathbf{f}_i, D^{(i)}, \epsilon$)
\EndFor
\State  \Return $\mathcal{T}$
\end{algorithmic}
\textbf{subroutine MinimizeDL}(transformation, $\mathbf{f}_i,D^{(i)},\ell,\mathcal{L}$)

\begin{algorithmic}[1]
\algrenewcommand{\alglinenumber}[1]{\bfseries s{#1}:}
\renewcommand{\Statex}{\item[\hphantom{\bfseries Step \arabic{ALG@line}.}]}

\While {transformation.is\_applicable($\mathbf{f}_i$)}
\State dl$_0$ $\leftarrow$ DL$(\mathbf{f}_i)$ + $\sum_{(\mathbf{x}_t,\mathbf{y}_t)\in D^{(i)}} \ell_{DL,\epsilon} [\mathbf{f}_i(\mathbf{x}_t),\mathbf{y}_t]$
\State $f_{\mathsf{clone}} \leftarrow \mathbf{f}_i$ // \textit{clone $\textbf{f}_i$ in case transformation fails}
\State $\mathbf{f}_i\leftarrow $ transformation($\mathbf{f}_i$)
\State $\mathbf{f}_i\leftarrow $ Minimize$_{\mathbf{f}_i}$$\sum_{(\mathbf{x}_t,\mathbf{y}_t)\in D^{(i)}} \ell_{DL,\epsilon} [\mathbf{f}_i(\mathbf{x}_t),\mathbf{y}_t]$
\State dl$_1$ $\leftarrow$ DL$(\mathbf{f}_i)$ + $\sum_{(\mathbf{x}_t,\mathbf{y}_t)\in D^{(i)}} \ell_{DL,\epsilon} [\mathbf{f}_i(\mathbf{x}_t),\mathbf{y}_t]$
\State \textbf{if} dl$_1>$ dl$_0$ \textbf{return} $f_{\mathsf{clone}}$
\EndWhile
\State \Return $\mathbf{f}_i$
\end{algorithmic}
\end{algorithm}

The rationalSnap transformation (step 6) replaces a real or an irrational parameter in $f_i$ by a rational number. This significantly reduces the $DL(f_\theta)$. For example, if we have $\pi=3.14159265359...$, the algorithm (step 6) will replace it by $p={355\over 113}$ and then calculate the DL. It will find out that this snap minimized the total DL. To calculate the description length of a rational number that has an integer $m$ as a numerator and a natural number $n$ as a denominator we use:
\begin{equation}
DL({m\over n})=\log_2[(1+|m|)n].
\end{equation}
Next comes the toSymbolic (step 7) that transforms the prediction functions into a symbolic representation.

After defining these transformations, they are fed to the subroutine MinimizeDL function as variables in addition to the $D^{(i)}$'s and $\epsilon$. The subroutine takes $f_i$, $D^{(i)}$, and the transformation as input and repeatedly applies the transformation to $f_i$. The subroutine MinimizeDL function starts calculating the dl of the prediction function (s2), then duplicates the prediction function $f_{clone}$ (s3) and stores it to be reused in the case where the transformation fails. In steps (s4-s5), the algorithm starts by performing the transformation and takes into account minimizing the loss. The transformation is accepted if the description length ``dl" of the theory $i$ decreased, with a $0$-step patience implementation, where 
 \begin{equation*}
     dl=DL(f_i)+\sum_{(x_t,y_t)\in D^{(i)}}l_{DL,\epsilon}(f_i(x_t),y_t).
 \end{equation*} 
which means if the new description length has increased, directly exit the loop and take the prediction function before the transformation or $f_{clone}$ (step s7). If the transformation keeps the description length unchanged, redo the transformation with a $4$-step patience; if the ``dl" remain unchanged then exit the loop.


\subsubsection{Unification algorithm}
Towards the ultimate goal of science, which is the simulation of nature, the idea is not just describing the phenomena observed, but also to seek any connections between them. This helps to unify theories behind these phenomena. The following algorithm will show how the unification process takes place in the "AI Physicist".

The algorithm will take the symbolic prediction function $\{(f_i,.)\}$ and outputs master theories $\mathcal{T}=\{(f_p,.)\}$. By modifying $p$ in $f_p$, we can generate a continuum of already known prediction functions $f_i$. The talk now will be on how to output or find a master theory.
\begin{algorithm}[H]
\caption{\textbf{Theory Unification}}
\begin{algorithmic}[1] 
\label{unification}
\Require \textbf{Hub}: theory hub
\Require $K$: initial number of clusters
\For{$(\mathbf{f}_i, c_i)$ \textbf{in Hub}.all-symbolic-theories}
\State dl$^{(i)}$ $\leftarrow$ DL($\mathbf{f}_i$)
\EndFor
\State  $\{S_k\} \leftarrow$ Cluster ($\mathbf{f}_i$) into $K$ clusters based on dl$^{(i)}$

\For {$S_k$ \textbf{in} $\{S_k\}$}
\State  $\left(\mathbf{g}_{ik},\mathbf{h}_{ik}\right) \leftarrow$ \textbf{Canonicalize}($\mathbf{f}_i$), $\forall \mathbf{f}_i \in S_k$
\State  $\mathbf{h}_k^* \leftarrow$ Mode of $\{ \mathbf{h}_{ik}|\mathbf{f}_{ik}\in S_k \}.$
\State  $G_k \leftarrow \{ \mathbf{g}_{ik}|\mathbf{h}_{ik}=\mathbf{h}_k^* \}$
\State  $\mathbf{g}_{\mathbf{p}k} \leftarrow$ Traverse all $\mathbf{g}_{ik}\in G_k$ with synchronized steps,
replacing the coefficient by a $\mathbf{p}_{jk}$ when not all
   coefficients at the same position are identical.
\State  $\mathbf{f}_{\mathbf{p}k} \leftarrow$ toPlainForm$(\mathbf{g}_{\mathbf{p}k})$
\EndFor
\State $\mathscr{T} \leftarrow \{(\mathbf{f}_{\mathbf{p}k},\cdot)\}$, $k=1,2,\dots K$
\State $\mathscr{T} \leftarrow$ MergeSameForm( $\mathscr{T}$)
\State  \Return $\mathcal{T}$
\newline

\end{algorithmic}
\textbf{subroutine Canonicalize}($\mathbf{f}_i$):
\begin{algorithmic}[1]
\algrenewcommand{\alglinenumber}[1]{\bfseries s{#1}:}
\renewcommand{\Statex}{\item[\hphantom{\bfseries Step \arabic{ALG@line}.}]}

\State $\mathbf{g}_i \leftarrow$ ToTreeForm$(\mathbf{f}_i)$
\State $\mathbf{h}_i \leftarrow$ Replace all non-input coefficient by a symbol $s$\newline
\Return $(\mathbf{g}_i,\mathbf{h}_i)$
\end{algorithmic}
\end{algorithm}

The algorithm takes the symbolic prediction functions as inputs, then it takes each of these functions and calculates its description length $dl^{(i)}$ (step 1-3). In (step 4), the algorithm clusters the symbolic functions $f_i$ based on their description length using, for example, $K$-mean clustering Appendix (\ref{k-mean}). The prediction functions thus become additionally labeled by the cluster number $k$: $f_{ik}$.
The unification process begins in (step 5-11). First, it starts by transforming the prediction function $f_{ik}$ into a 2-tuple $(g_{ik},h_{ik})$ (step 6), where $g_{ik}$ is the Tree-form of $f_{ik}$ and $h_{ik}$ is the structure of the tree. This transformation is performed by applying the ``subroutine Canonicalize'' function (step s1-s2). The steps 7 and 8 will find the trees that have the same structure $\textbf{\textit{h}}$ and group them in $G_{k}$. The coefficients that differ in the trees within the same group $G_k$ are parameterized by $p_{jk}$ (step 9). This parametrization has unified the trees into a master tree parametrized by $\textit{\textbf{p}}$. Once a master tree is found, the algorithm will re-transform it to the symbolic form (step 10) which is the master prediction function. Finally, the algorithm will update the master functions in $\mathcal{T}=\{(f_{p_k},.)\}$ and returns $\mathcal{T}$.


\subsection{Summing up Acquired Knowledge}
To have a better idea on how these algorithms work together, the data given to the ``AI Physicist" must be specified. As mentioned, the data is a time-series of vectors, each describing a two-dimensional motion of balls wandering four different domains. Each domain is characterized by a physical effect, either gravity, springs, electromagnetic field, or bounce boundaries. The goal is to predict the two-dimensional motion of these particles as accurately as possible. The ``AI physicist" starts by proposing theories from the theory hub, and it tests if they fit well a part of the data. Having the proposed theories and those initialized randomly, the agent proceeds to the divide-and-conquer algorithm. From the results of Tegmark et al in \citep{wu2018toward}, the ``AI Physicist" is able to construct the prediction function for predicting the ball's future positions and simultaneously classify them into the four aforementioned domains. The constructed prediction functions, in hand, are passed next to Occam's Razor to minimize their description length. Upon applying these applications, one of the difference equations, i.e. the prediction function, belonging to one of the four domains is executed as follows
\begin{equation}
\begin{aligned}
	 \mathbf{\hat y}_t=&\begin{pmatrix}
		-0,99999994 & 0.00000006 & 1.99999990 & -0.00000012\\
		-0.00000004 & -1.0000000 & 0.00000004 & 2.00000000
	\end{pmatrix}\mathbf{x}_t
	\\
	&+\begin{pmatrix}
		 0.01088213\\
		 -0.00776199
	\end{pmatrix},
\end{aligned}
\end{equation}
with the description length of the prediction function $DL(f)=212.7$. It is worth mentioning that this equations results after applying the first collapse-Layer transformation. When applying then the snapping transformation, it is further simplified
\begin{equation}\label{Eq5.10}
	 \mathbf{\hat y}_t=\begin{pmatrix}
		-1 & 0 & 2 & 0\\
		-0 & -1 & 0 & 2
	\end{pmatrix}\mathbf{x}_t+
	\begin{pmatrix}
		 0.010882\\
		 -0.007762
	\end{pmatrix},
\end{equation}
with a lower description length $DL(f)=55.6$. Still within the Occam's Razor strategy, the toSymbolic transformation transforms Eq.\ref{Eq5.10} into a symbolic expression
\begin{equation}
   \begin{aligned}
	&\hat x_{t+2}=2x_{t+1}- x_t +0.010882,\\
	&\hat y_{t+2}=2y_{t+1}- y_t -0.007762.
	\label{exam}
\end{aligned} 
\end{equation}

What remains now is to find if there is a master theory that groups the prediction functions talking about one theme. For that, the AI Physicist uses the unification algorithm to cluster the symbolic prediction functions based on their description length where the clusters have DL that ranges between certain thresholds. Finally, Eq.(\ref{exam}) is unified into a master theory $f_p$ with:
\begin{equation}
\begin{aligned}
	&\hat x_{t+2}=2x_{t+1}- x_t +p_1,\\
	&\hat y_{t+2}=2y_{t+1}- y_t +p_2.
	\label{symp}
\end{aligned}
\end{equation}
Based on these equations, it can be inferred that balls in this region are affected by gravity. To be clearer, Eq.(\ref{symp}) is rewritten as
\begin{equation}
\begin{aligned}
	&\ddot{x}=g_x,\\
	&\ddot{y}=g_y,
\end{aligned}
\end{equation}
where $g_i \equiv p_i(\Delta t)^2$. Concerning the other three domains, the one affected by springs, the one in electromagnetic field, and the final one with bounce boundaries, they are inferred in the same way this domain is deduced.

The results the ``AI Physicist" achieved are very promising to the extent that it outperforms alternative neural networks having approximately the same complexity. It is worth mentioning how important the lifelong learning strategy is within the learning algorithm. To check this, a new learning agent is constructed and named as ``Newborn AI Physicist". The main particularity of this new agent is that it has an empty theory hub. This means that it does not have any previously learned theories saved. This is a main difference between the ``Newborn AI Physicist" and the ``AI Physicist". Both achieved perfect accuracy levels but the ``AI Physicist" is able to learn faster than the other. This means that using previously learned theories helped in increasing the learning speed.

\subsubsection{Eliminating transition domains}

After having successfully described the theory governing each domain, the "AI Physicist" faces another problem which is the boundaries and transition regions. Which theory governs the motion when the trajectory moves from one domain to another each of different physics? Our AI agent typically finds the next position from the last $T = 3$ ones $\textbf{x}_t = (\textbf{y}_{t-3}, \textbf{y}_{t-2}, \textbf{y}_{t-1})$. When the trajectory is near a boundary, the set of previous positions can contain 3 positions from the $1^{st}$ domain and none from the $2^{nd}$, 2 positions from the $1^{st}$ and 1 from the $2^{nd}$, or vice versa. Each of these four situations require a different function to compute the next position. These situations increase in number as more domains are added.

Moreover, the domains are set to be perfectly elastic. That is, if the trajectory encounters a boundary (wall) of one domain, it will be reflected back (bounces). These situations also require different functions to govern them. For transitions and bounces, the data is insufficient and thus, the agent will fail to find solutions. However, a straightforward solution to this problem is to simply eliminate these regions using the following steps:
\begin{enumerate}
\item For each domain where the function $\textbf{f}$ that predicts future positions is found ($\textbf{x}_t \rightarrow \textbf{y}_{t}$) , we find the function that predicts the past positions ($\textbf{x}_t \rightarrow \textbf{y}_{t-T-1}$).
\item When a trajectory approaches a transition region and using the future-prediction function of the first domain, an extrapolation that is forward in time is performed. The extrapolation is a forward extension of the trajectory as if it is still in the first domain. We fit the function $\textbf{y}_f(t)$ for this forward extrapolation.
\item For the same trajectory and using the past-prediction function of the second domain, an etxrapolation that is backwards in time is performed. This extrapolation is a backward extension of trajectory as if it was in the second domain. We fit the function $\textbf{y}_b(t)$ for this backward extrapolation.
\item We find the time $t_*$ where the difference between these functions is minimized.
\begin{equation}
t_* = \arg \min_{t} |\mathbf{y}_f(t) - \mathbf{y}_b(t)|.
\end{equation}
In the ideal case, it will be exactly the intersection point of these two functions.
\end{enumerate}
If at this point we have $\textbf{y}_f(t_*) \approx \textbf{y}_b(t_*)$, then it is a boundary point. Additionally, if $\textbf{y}'_f(t_*) \approx \textbf{y}'_b(t_*)$, then we're dealing with the transition boundary between one region and another since the near this boundary, the speeds don't differ upon transition. Otherwise, the boundary will be an exterior one (bouncing). Once these points are found, the domain classifiers are trained again to eliminate these points and extend the domains as much as possible.


\section{CONCLUSION}


\subsection{Concluding Remarks}
It is hoped that this review contributes to give an obvious insight of topics at the intersection of both scientific fields: artificial intelligence and physics. In this regard, both fields played jointly an important role in making significant advances. This review introduces first an overview of neural networks towards understanding how they are adopted in algorithms and how they are efficiently trained using different types of learning paradigms. In particular, one of these paradigms is reinforcement learning which is concerned in modeling an agent's performance in stochastic environments. Problems integrating reinforcement learning techniques acquire significant interest because they embody real world situations characterized by stochastic behaviors. These problems are described by the so called Markov decision processes.  The interest arising in neural networks is their usage in algorithms to help improve the understanding of physical theories behind observational data. One of them is \textit{SciNet}, a neural network constructed by Renato \textit{et al} in \citep{iten2018discovering}. Its main role is to infer a minimal representation of observational data that summarizes all the important aspects of their physical setting.  Another technique, the ``AI Physicist" created by Tegmark \textit{et al} in \citep{wu2018toward}, incarnates the physicist's way of thinking and encountering problems. The technique is divided into four strategies summarizing a physicist's way of tackling problems. These mentioned techniques show promising results. Physicists working towards this direction of research can rely on those results and make any advances extending the current work. Many questions are still open to future research work, some of which are mentioned in the next section.


\subsection{Outlook}
The tools of artificial intelligence contribute deeply into solving physical problems and further revealing the intuition behind them. However, the concepts of physics can be embedded to strengthen AI as well. Renormalization group can be illustrated in deep neural networks as a scheme for extracting features from data \citep{mehta2014exact,li2018neural}. Going beyond the classical domain, quantum coherence and entanglement are used in quantum computers making them more efficient than classical ones. These techniques can be implemented in machine learning \citep{deng2017quantum} thus speeding up data processing, but quantum machine learning still suffers from high cost and complexity \citep{biamonte2017quantum}. However,  tools such as symmetries and gauge fixing may maximize the powerfulness of the ``AI Physicist" program and constrain the remaining problems in such a way that will enable us to lower the complexity and reduce the effect of the expected noise. These issues and some related problems are currently under investigation.

The conventional ``AI Physicist" studied in this review article has clearly made use of classical computer. Recently, several serious attempts have been put forward to extend machine learning into the quantum settings \citep{biamonte2017quantum,scholkopf2002learning,wittek2014quantum,schuld2018supervised}. Surprisingly, machine learning, being based on Kernel methods \citep{scholkopf2002learning} (e.g., vector machines (SVMs)), turns out to share some similar theoretical foundations with quantum computation in that it efficiently puts in effect computation in arbitrarily large Hilbert space. Such a connection paves the road for the experts to design quantum machine learning algorithms (see \citep{wiebe2012quantum,schuld2019quantum,havlivcek2019supervised} and references therein). To this end, a general approach to build a quantum neural network has been introduced in \citep{killoran2018continuous}. Thus, a natural extension of the actual work is to upgrade the ``AI Physicist" to a quantum setup ushering us into the era of the \emph{quantum AI Physicist} \citep{we}.

A related link which has recently gained a considerable momentum is machine learning with quantum-inspired tensor networks (see \citep{huggins2018towards, han2018unsupervised} and references therein). Tensor networks can be exploited in adaptive and unsupervised learning analogous to renormalization group (RG). The tantalizing part about this result is the developed framework which has the advantage of unifying the designed algorithms and theoretical developments in a way that it is prolific for both classical and quantum computing. In such a unified scheme, one ought to train the model by an ``AI Physicist" then pass it on to a  quantum AI Physicist for further optimization. It would be beneficial to examine this connection further and apply it to problems in condensed matter and high energy physics (e.g, AdS/CFT duality \citep{you2018machine,hashimoto2018deep}). 

Moreover, employing the SciNet's encoder in ``AI Physicist"'s  Occam's razor in order to extract the minimal representation can be a step towards lowering the training time of computations even further. This would require a modification in the ``AI Physicist" architecture which could lead to an improvement in performance.


\begin{acknowledgements}
This review article is based on a series of lectures and seminars given at the Lebanese University during the fall of 2018 and spring 2019. The authors would like to thank the Lebanese University for providing the necessary resources that enabled the authors to complete this work. W.A.C would like to thank the Institute for Quantum Information and Matter (IQIM), Caltech, for the ongoing stimulating environment from which the author has been significantly benefited. W. A.C gratefully acknowledges the support of the Natural Sciences and Engineering Research Council of Canada (NSERC). 
\end{acknowledgements}

\clearpage


\begin{appendix}
\section{Activation Functions}\label{activation}
The following table presents a list of the commonly used activation functions. It includes the definition of each one with its properties.

 \begin{figure}[h]
     \centering
     \includegraphics[totalheight=11.2cm]{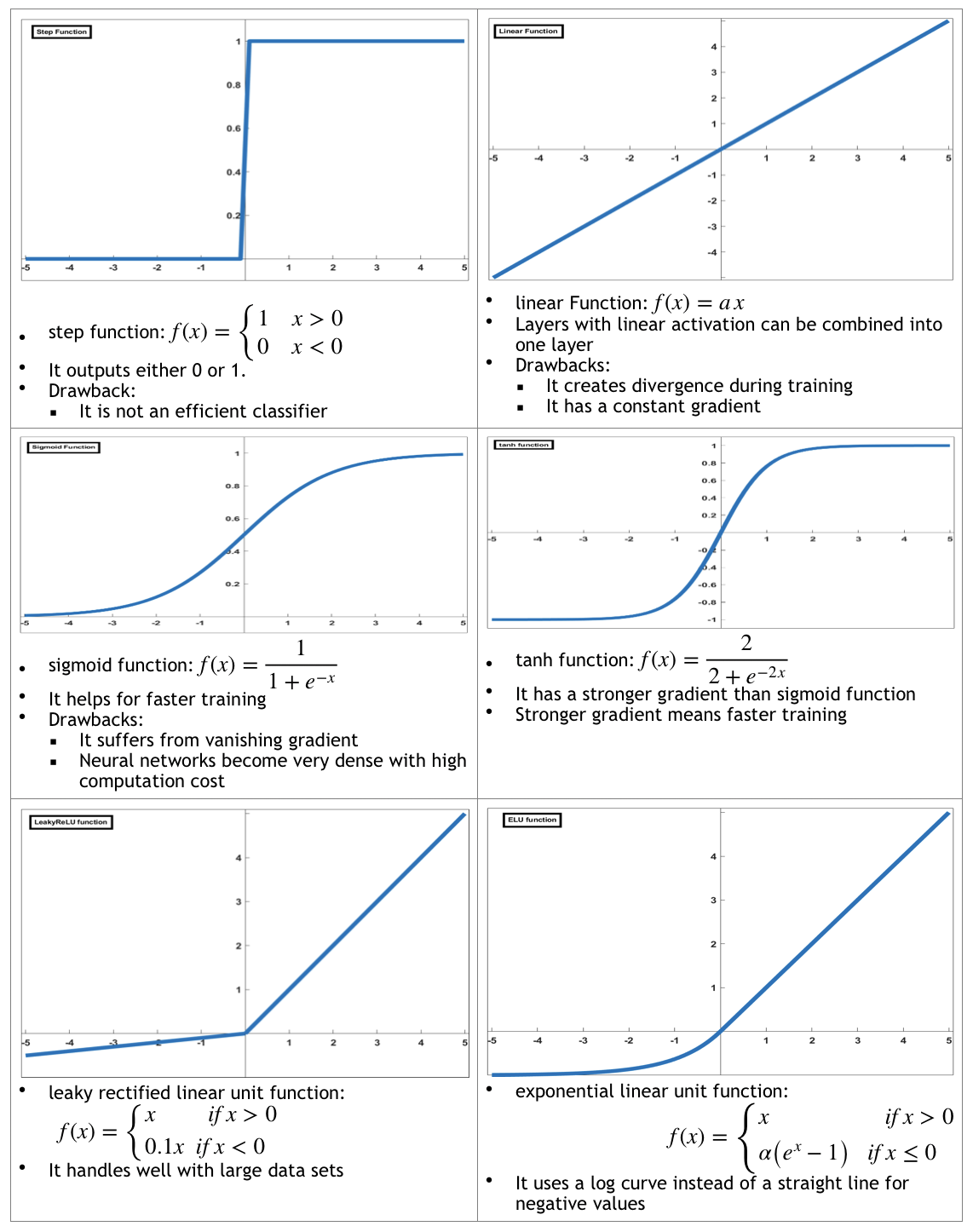}
 \end{figure}  

\section{K-means Clustering}\label{k-mean}
The objective of $K$-means clustering \citep{macqueen1967} is simple: group similar data points together and discover underlying patterns. The algorithm looks for a fixed number $K$ of clusters in a dataset. A cluster is a collection of data points grouped together using a distance metric with a certain threshold, and the number of clusters is initialized depending on one's choice. The idea is as follows: suppose having a huge dataset where training this set all together is very tedious and costly. The $K$-mean clustering algorithm is used to divide the set into subsets where it is easier to solve each cluster on its own. The algorithm converges to a solution after some iterations summarized in four steps:
\begin{enumerate}
	\item Initialize the centers (centroids) by choosing $K$ data points at random. 
	\item Assign data points to clusters by calculating the distance between the data points and the centroids.
	\item Update clusters centroids with new values which is the average or the mean of all data points within the cluster.
	\item Repeat till the desired one of these conditions is met:\begin{enumerate}
	\item The data points assigned to each cluster remain the same.
	\item Centroids becomes fixed through repetitive iterations.
	\item The distance between the data points and the centroids is minimum.
	\item The number of iterations should be sufficient to guarantee convergence.
	    \end{enumerate}
\end{enumerate}

 \clearpage
 
\end{appendix}

\clearpage

\begin{figure}

\begin{subfigure}{1\textwidth}
        \includegraphics[width=\textwidth,height=15cm]{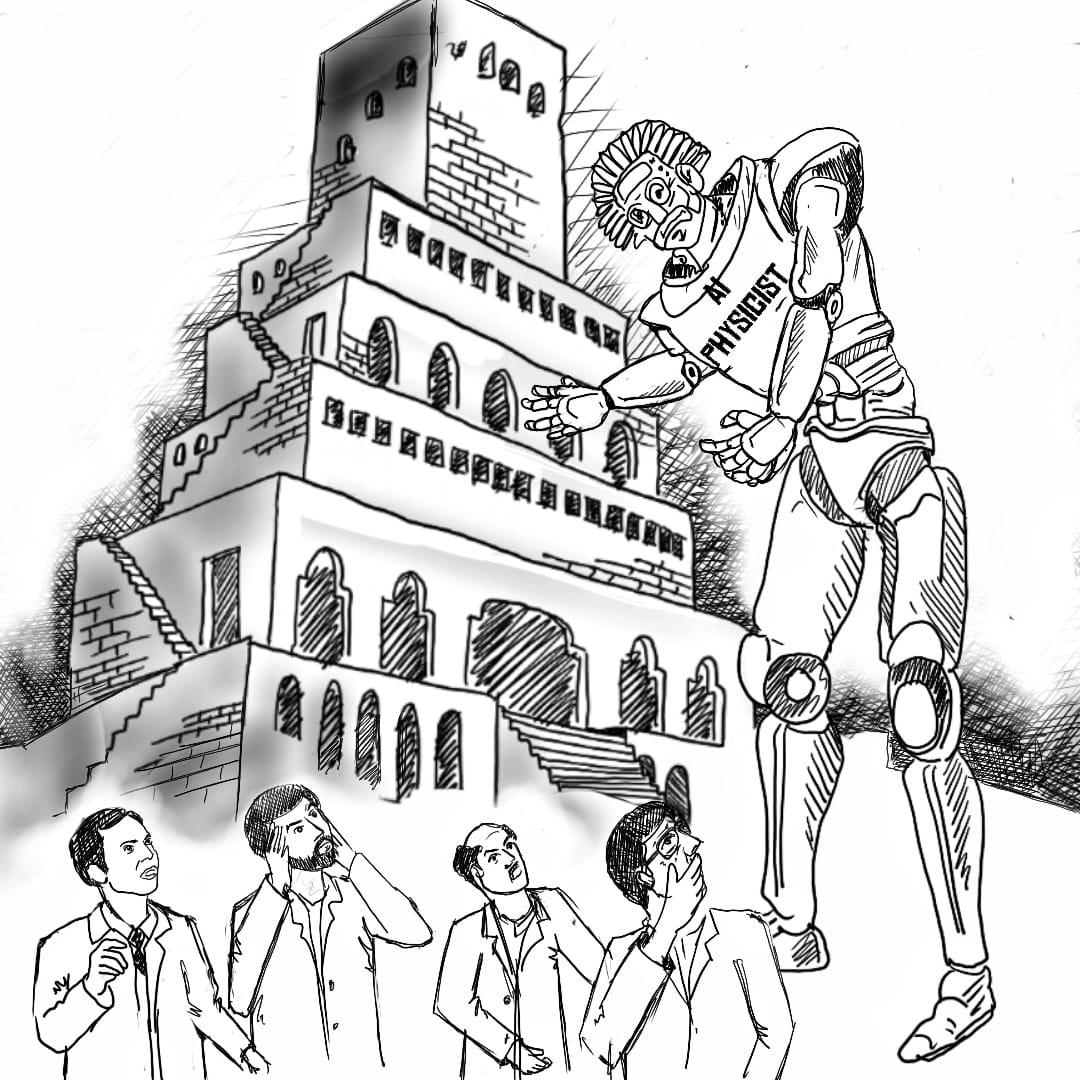}
        \caption*{\textit{AI Physicist, The Conqueror of Babel}}
    \end{subfigure}
\end{figure}

%

\end{document}